\documentclass[sigconf,nonacm]{acmart}
\AtBeginDocument{%
  }

\usepackage{amsthm}
\usepackage{amsfonts}
\usepackage{caption}
\usepackage{subcaption}
\usepackage{multirow}
\usepackage{rotating}
\usepackage{amsmath}
\usepackage{graphicx}
\usepackage{booktabs}
\usepackage{enumitem}
\usepackage{makecell}

\begin{document}

\title{Learning from Points and Patches: A Pre-trained Hybrid Attention Transformer for Time Series Analysis}

\author{Peiwang Tang}
\affiliation{%
  \institution{University of Science and Technology of China}
  \city{Hefei}
  \state{Anhui}
  \country{China}
}
\affiliation{%
  \institution{State Key Laboratory of Cognitive Intelligence}
  \city{Hefei}
  \state{Anhui}
  \country{China}
}
\affiliation{%
  \institution{Shanghai Black Wing Asset Management}
  \city{Shanghai}
  \country{China}
}
\email{tpw@mail.ustc.edu.cn}

\author{Cheng HE}
\affiliation{%
  \institution{University of Science and Technology of China}
  \city{Hefei}
  \state{Anhui}
  \country{China}
}
\affiliation{%
  \institution{State Key Laboratory of Cognitive Intelligence}
  \city{Hefei}
  \state{Anhui}
  \country{China}
}
\affiliation{%
  \institution{Shanghai Black Wing Asset Management}
  \city{Shanghai}
  \country{China}
}
\email{cheng.he@mail.ustc.edu.cn}

\author{Yupeng Ye}
\affiliation{%
  \institution{University of Science and Technology of China}
  \city{Hefei}
  \state{Anhui}
  \country{China}
}
\email{yeyupeng@mail.ustc.edu.cn}

\author{Weitai Zhang}
\affiliation{%
  \institution{iFLYTEK Research}
  \city{Hefei}
  \state{Anhui}
  \country{China}
}
\email{zwt2021@mail.ustc.edu.cn}

\author{Qinghua Sun}
\affiliation{%
  \institution{iFLYTEK Research}
  \city{Hefei}
  \state{Anhui}
  \country{China}
}
\email{qhsun2@iflytek.com}

\author{Defu Lian}
\affiliation{%
  \institution{University of Science and Technology of China}
  \city{Hefei}
  \state{Anhui}
  \country{China}
}
\affiliation{%
  \institution{State Key Laboratory of Cognitive Intelligence}
  \city{Hefei}
  \state{Anhui}
  \country{China}
}
\email{liandefu@ustc.edu.cn}

\renewcommand{\shortauthors}{Tang et al.}

\makeatletter
\renewcommand{\@mkauthors@iii}{%
  \gdef\@currentauthors{}%
  \gdef\@currentaffiliation{}%
  \global\setbox\mktitle@bx=\vbox{%
    \unvbox\mktitle@bx\par\medskip
    \begingroup
    \centering\normalfont
    {\fontsize{11}{14}\selectfont
      Peiwang Tang\textsuperscript{1,2,3}\quad
      Cheng HE\textsuperscript{1,2,3}\quad
      Yupeng Ye\textsuperscript{1}\quad
      Weitai Zhang\textsuperscript{4}\quad
      Qinghua Sun\textsuperscript{4}\quad
      Defu Lian\textsuperscript{1,2,\ding{41}}\par}
    \smallskip
    {\fontsize{9}{12}\selectfont
      \textsuperscript{1}University of Science and Technology of China, Hefei, Anhui, China\par
      \textsuperscript{2}State Key Laboratory of Cognitive Intelligence, Hefei, Anhui, China\par
      \textsuperscript{3}Shanghai Black Wing Asset Management, Shanghai, China\par
      \textsuperscript{4}iFLYTEK Research, Hefei, Anhui, China\par}
    \smallskip
    {\fontsize{8}{11}\selectfont
      \href{mailto:tpw@mail.ustc.edu.cn}{\texttt{tpw@mail.ustc.edu.cn}}\quad
      \href{mailto:cheng.he@mail.ustc.edu.cn}{\texttt{cheng.he@mail.ustc.edu.cn}}\quad
      \href{mailto:yeyupeng@mail.ustc.edu.cn}{\texttt{yeyupeng@mail.ustc.edu.cn}}\par
      \href{mailto:zwt2021@mail.ustc.edu.cn}{\texttt{zwt2021@mail.ustc.edu.cn}}\quad
      \href{mailto:qhsun2@iflytek.com}{\texttt{qhsun2@iflytek.com}}\quad
      \textsuperscript{\ding{41}}\href{mailto:liandefu@ustc.edu.cn}{\texttt{liandefu@ustc.edu.cn}}\par}
    \endgroup
    \bigskip
  }%
}
\makeatother

\begin{abstract}
Time-series data generated by Web services and connected systems support applications such as user activity analysis, service monitoring, and resource management. However, learning transferable representations from heterogeneous time series remains challenging due to diverse temporal patterns, variable sequence lengths, and limited high-quality annotations. Despite the success of pre-training in natural language processing and computer vision, effective time-series pre-training requires strategies tailored to temporal data.
We propose TimeHAT (Hybrid Attention Transformer), a general-purpose pre-training framework for time-series analysis that addresses two complementary questions: what to learn and how to learn. For what to learn, we introduce a Unified Point-Patch Masking Strategy (UPPMS) that integrates point-level and patch-level supervision within a unified framework for contextual reconstruction and causal extrapolation, enabling the model to capture both fine-grained temporal variations and broader temporal patterns. For how to learn, we develop a Hybrid Attention Encoder (HAE) that combines prefix-constrained bidirectional attention in initial layers with directional hybrid attention in deeper layers, aiming to preserve effective contextual interactions while mitigating rank degeneration.
We evaluate TimeHAT across real-world datasets and four downstream tasks: classification, long-term forecasting, imputation, and anomaly detection. Experiments cover fine-tuning, few-shot, and zero-shot transfer settings, demonstrating competitive performance across diverse time-series analysis tasks.
Code is available at \url{https://github.com/TangPeiwang/TimeHAT}.
\end{abstract}



\maketitle

\section{Introduction}

\begin{figure}[ht]
	\centering
	\includegraphics[scale=0.18]{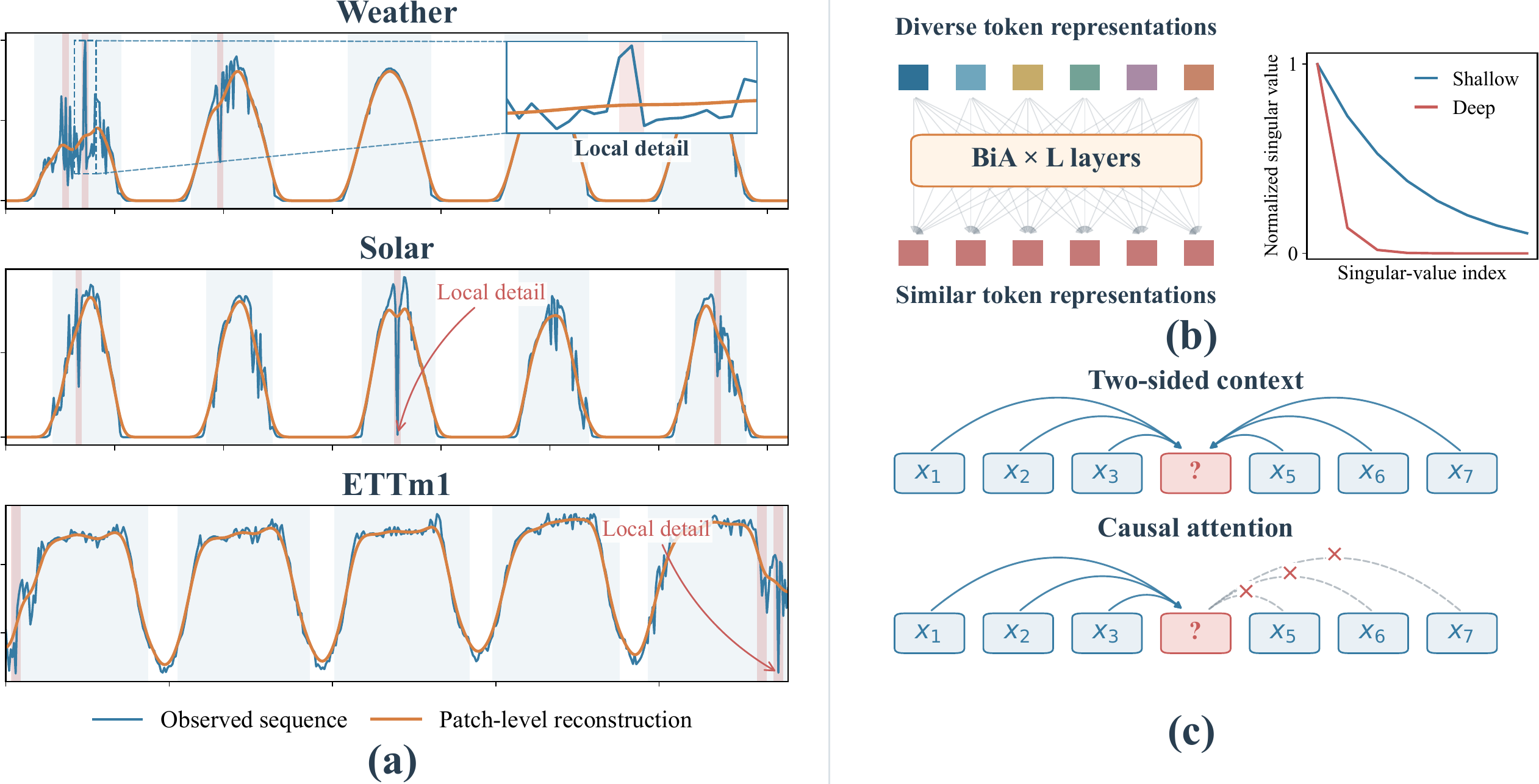}
  \caption{(a) Patch-level reconstruction can miss fine-grained temporal details;
(b)	Deep bidirectional attention risks representation homogenization;
(c)	Causal attention cannot access two-sided context.}
	\label{fig:motivation}
\end{figure}

The World Wide Web continuously generates time-series signals through user interactions, network traffic, and service telemetry \cite{zhang2026semixer, zhang2026re, zhang2026time}. These signals underpin a wide range of Web applications, including sequential recommendation, user behavior modeling, traffic forecasting, resource management, and service anomaly detection \cite{wang2026eeo, zhang2026enhancing, zhang2026amortized}. Beyond Web systems, time-series data (TSD), collected across sequential time points to monitor dynamic phenomena, serve as a cornerstone for decision-making in diverse scientific and industrial domains \cite{liu2024unitime,xie2025multivariate}. Despite their broad utility, real-world TSD are often heterogeneous across domains, exhibit diverse temporal patterns and pervasive noise, and frequently lack high-quality annotations \cite{huang2025exploiting}. These challenges motivate the development of general-purpose pre-training methods that can learn transferable representations from diverse time series and support a broad range of downstream tasks, including forecasting, classification, imputation, and anomaly detection.

\begin{figure*}[ht]
	\centering
	\includegraphics[scale=0.56]{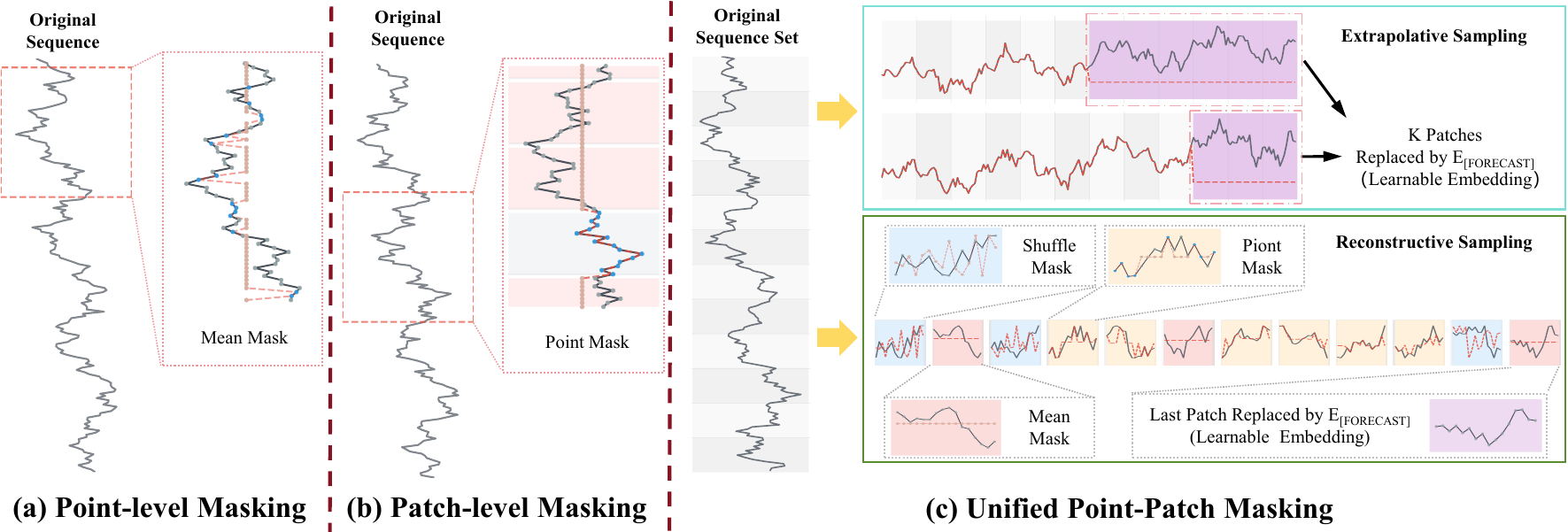}
  \caption{Comparison of Point-Level, Patch-Level, and Unified Point-Patch Masking Strategies for Time Series Data.}
	\label{masked}
\end{figure*}

A natural approach to learning such representations is to draw inspiration from the large-scale pre-training paradigm that has achieved remarkable success in natural language processing (NLP) \cite{devlin2018bert} and computer vision (CV) \cite{he2022masked}. However, directly transferring these paradigms to time-series analysis (TSA) remains challenging \cite{liu2023large,gruver2023large}. Unlike NLP and CV, which benefit from large-scale corpora and established pre-training resources such as ImageNet \cite{zhou2023one}, TSA faces substantial data scarcity and heterogeneity across datasets and domains. Consequently, existing pre-training strategies may fail to adequately capture the distinctive statistical properties of time series \cite{sun2023test,chang2025llm4ts}. This mismatch suggests that effective time-series pre-training requires not only sufficient data and model capacity, but also learning strategies and architectures tailored to temporal data \cite{liu2024timer,gao2024units}.

Despite recent progress, existing time-series pre-training methods face
limitations in both reconstruction granularity and contextual modeling.
(As illustrated in Figure~\ref{fig:motivation}). These limitations
motivate two fundamental questions: \textbf{what to learn} from temporal
data and \textbf{how to learn} it effectively.

Existing masking-based methods primarily reconstruct masked patches,
providing supervision for segment-level temporal patterns
\cite{tang2022mtsmae,nie2022time}. However, patch-level reconstruction
does not explicitly supervise fine-grained variations within each patch,
which may also carry important temporal information. Conversely,
point-level masking provides finer-grained supervision but does not
explicitly capture segment-level structure. Effective pre-training should
therefore capture complementary temporal information at both point and
patch granularities.

However, determining what temporal information to learn is only part of
the challenge. The encoder must also integrate contextual information
while preserving diverse representations across layers. Standard
bidirectional attention (BiA), widely used in masked reconstruction,
enables global token interactions but can suffer from representation
homogenization and rank degeneration in deep networks
\cite{bhojanapalli2020low,dong2021attention}. Directional attention can
mitigate this degeneration, but applying directional masks from the
initial layer may limit early global contextual competition. This
motivates an encoder that combines global contextual integration in
shallow layers with directionally structured attention in deeper layers.

To address these challenges, we propose TimeHAT (Hybrid Attention
Transformer), a general-purpose pre-training framework for time-series
analysis. For \textit{what to learn}, we introduce a Unified Point-Patch
Masking Strategy (UPPMS) that unifies contextual reconstruction and
causal extrapolation, using point-level and patch-level supervision to
capture complementary temporal information. For \textit{how to learn},
we develop a Hybrid Attention Encoder (HAE) with a progressive attention
architecture. It first employs prefix-constrained bidirectional
attention to establish contextual representations, then introduces
complementary forward and backward attention heads in deeper layers
to mitigate rank degeneration while preserving causal constraints on
predictive targets. We evaluate TimeHAT across diverse downstream tasks
and transfer settings, and further investigate its effectiveness under
limited-data conditions.

Our contributions are summarized as follows:
\begin{itemize}
    \item We propose UPPMS, a unified masking framework that combines contextual
    reconstruction and causal extrapolation, incorporating point-level
    and patch-level supervision to capture complementary temporal
    information.

    \item We develop HAE, which combines prefix-constrained bidirectional
    attention in shallow layers with complementary directional attention
    in deeper layers, balancing contextual integration with rank stability.

    \item We evaluate TimeHAT across classification, long-term forecasting,
    imputation, and anomaly detection, including fine-tuning, few-shot,
    and zero-shot transfer settings. We further analyze its effectiveness
    under limited-data conditions.
\end{itemize}

\begin{figure*}[ht]
	\centering
	\includegraphics[scale=0.58]{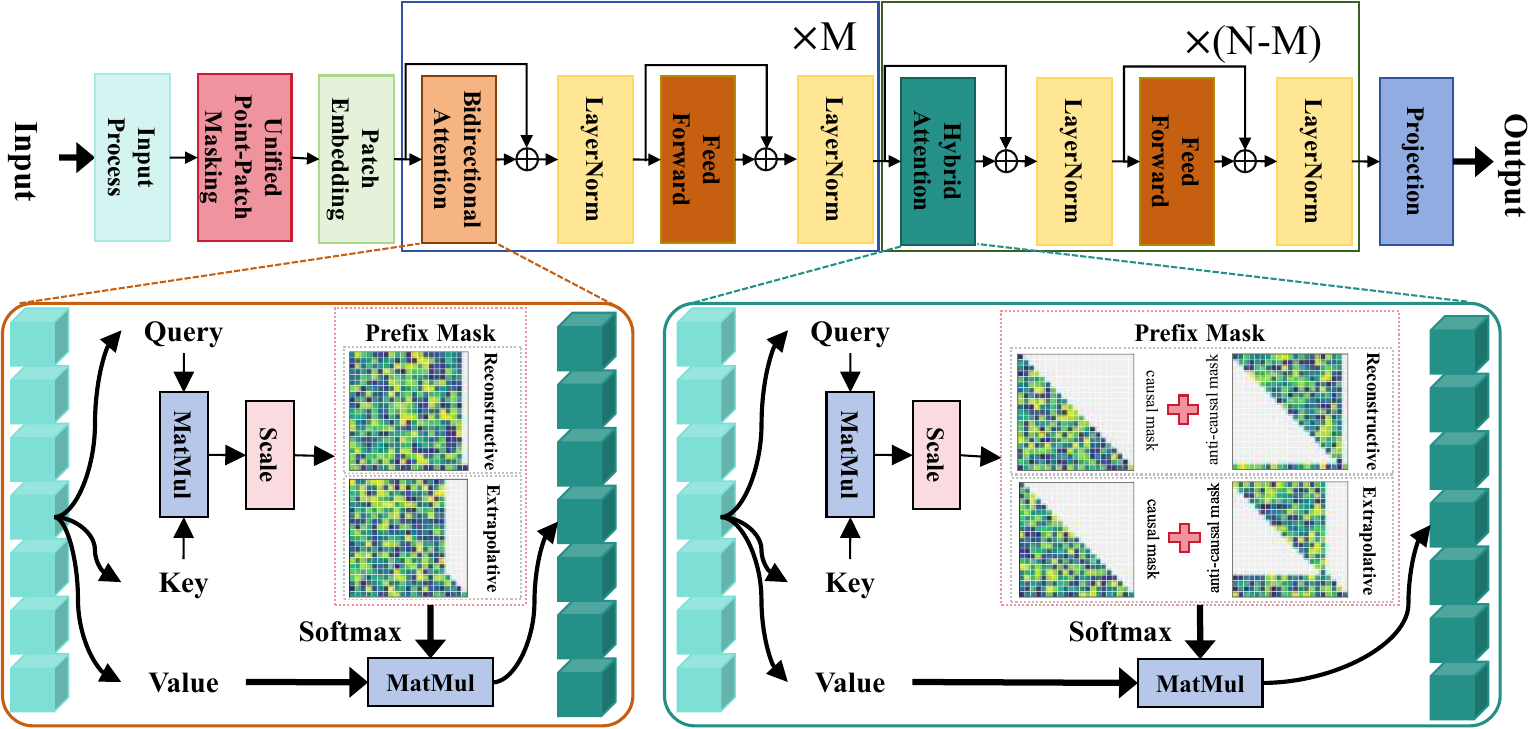}
  \caption{Overall architecture of TimeHAT. The model employs channel-independence for multivariate inputs. Sequences undergo a Unified Point-Patch Masking Strategy, patch tokenization, and MLP embedding. A two-stage encoder then captures temporal dynamics, utilizing Bidirectional Attention in early layers for global context, and Hybrid Attention in later layers for causal constraints. Finally, an MLP projection head reconstructs the output by modeling both patch- and point-level interactions.}
	\label{model}
\end{figure*}

\section{Unified Point-Patch Masking Strategy}

Recent masked time-series pre-training methods commonly reconstruct masked
patches \cite{tang2022mtsmae,nie2022time}. While patch-level masking
provides supervision for segment-level patterns, it does not explicitly
supervise fine-grained variations within each patch. Combining point-level
and patch-level masking can therefore provide complementary supervision
for temporal representation learning.

To address this limitation, we propose a Unified Point-Patch Masking
Strategy (UPPMS) that integrates point-level and patch-level masking
within a unified pre-training framework (\ref{masked}). UPPMS further supports two
complementary learning modes: contextual reconstruction from observed
sequences and causal extrapolation toward future segments.

Given a univariate time series $\mathbf{x}=\{x_1,\ldots,x_L\}$ of length
$L$, we divide it into $T=\lceil L/P\rceil$ non-overlapping patches
$\mathcal{P}=\{\mathbf{p}_1,\ldots,\mathbf{p}_T\}$, each of length $P$.
Zero-padding is applied to the final patch when necessary. 

\subsection{Dual-mode Sampling}

We define the perturbation set $\mathcal{I}_{\mathrm{pert}}$ and the
predictive target set $\mathcal{I}_{\mathrm{target}}$ to specify which
patches are corrupted and which serve as prediction targets. The masked
patch set is their disjoint union:
$
\mathcal{I}_{p}
=
\mathcal{I}_{\mathrm{pert}}
\cup
\mathcal{I}_{\mathrm{target}},
\qquad
\mathcal{I}_{\mathrm{pert}}
\cap
\mathcal{I}_{\mathrm{target}}
=
\emptyset.
$
During pre-training, UPPMS samples one of two modes according to a
predefined sampling probability:

\begin{itemize}
    \item \textbf{Contextual Reconstruction.}
    We uniformly sample $\lfloor r_p(T-1)\rfloor$ patches from the
    historical positions $\{1,\ldots,T-1\}$ to form
    $\mathcal{I}_{\mathrm{pert}}$, where $r_p\in(0,1)$ is the
    patch-level perturbation ratio. The final patch is designated as
    the prediction target:
    $
    \mathcal{I}_{\mathrm{target}}=\{T\}.
    $
    This mode reconstructs masked information from the historical context.
    The terminal target also serves as a \texttt{[CLS]}-like global
    readout for downstream tasks such as classification and anomaly
    detection.

    \item \textbf{Causal Extrapolation.}
    We disable historical perturbation,
    $\mathcal{I}_{\mathrm{pert}}=\emptyset$, and select a contiguous
    tail of $k$ patches as prediction targets:
    $
    \mathcal{I}_{\mathrm{target}}
    =
    \{T-k+1,\ldots,T\},
    $
    where $1\leq k\leq T$. This mode learns to predict future segments
    under causal information constraints.
\end{itemize}

\subsection{Masking Operations.}

\subsubsection{Predictive Target Masking.}
For each target patch $i\in\mathcal{I}_{\mathrm{target}}$, its projected
representation is replaced with a shared learnable query embedding
$\mathbf{E}_{\mathrm{forecast}}\in\mathbb{R}^{d_{\mathrm{model}}}$.
The query aggregates visible contextual information to reconstruct masked
content or predict future segments.

\subsubsection{Patch-level Masking.}
For each patch $i\in\mathcal{I}_{\mathrm{pert}}$, we randomly apply one
of three corruption operations according to predefined proportions:
(i) \textit{zero replacement}, which removes the patch content and
encourages recovery of missing segment-level structure;
(ii) \textit{mean replacement}, which replaces the patch with the global
scalar mean $\bar{x}_v$ of all visible observations, providing a
statistical reference while obscuring its original content; and
(iii) \textit{internal shuffling}, which randomly permutes the temporal
order within $\mathbf{p}_i$, encouraging the model to learn local
information beyond the original point ordering.

\subsubsection{Point-level Masking.}
Point-level masking is applied only in contextual reconstruction mode,
keeping the historical context unperturbed at the point level during
causal extrapolation. For each patch $i\notin\mathcal{I}_p$, we uniformly
sample a subset $\mathcal{I}_{pt,i}\subseteq\{1,\ldots,P\}$ of size
$\lfloor r_{pt}P\rfloor$, where $r_{pt}$ is the point-masking ratio.
Each selected value $p_{i,k}$, for $k\in\mathcal{I}_{pt,i}$, is replaced
with either zero or a sample from $\mathcal{N}(0,1)$. This operation
provides fine-grained reconstruction supervision within otherwise
visible patches.

\subsection{Prefix-constrained Attention.}
\label{sec:uppms}
The predictive target set $\mathcal{I}_{\mathrm{target}}$ defines a
prefix constraint mask $\mathbf{M}^{p}\in\mathbb{R}^{T\times T}$,
where $i$ and $j$ index the query and key positions, respectively:
\begin{equation}
M_{ij}^{p} =
\begin{cases}
-\infty, & i\notin\mathcal{I}_{\mathrm{target}},
             \ j\in\mathcal{I}_{\mathrm{target}},\\
-\infty, & i,j\in\mathcal{I}_{\mathrm{target}},\ j>i,\\
0, & \text{otherwise}.
\end{cases}
\end{equation}
This mask prevents historical queries from accessing target tokens
and enforces causal information flow among target tokens. In HAE,
it is combined with layer-specific attention masks to determine the
complete information flow, supporting both contextual reconstruction
and causal extrapolation without target leakage.

\section{TimeHAT}

We represent a multivariate time series as
$\mathbf{X}\in\mathbb{R}^{L\times C}$, where $L$ is the sequence length
and $C$ is the number of variables. TimeHAT adopts channel-independent
temporal encoding, applying the same operations independently to each
univariate series. This design focuses on learning temporal patterns
within individual variables without relying on potentially unstable
cross-variable dependencies.

For clarity, we describe the framework for a univariate series; the
multivariate extension applies the same encoder independently across
channels. As illustrated in Figure~\ref{model}, TimeHAT adopts an
encoder-only Transformer architecture inspired by masked pre-training
methods such as MAE. A lightweight linear projection maps encoder
representations to reconstruction targets, avoiding a separate Transformer
decoder and keeping the architecture focused on temporal representation
learning \cite{zeng2023transformers,chen2023tsmixer}.

\subsection{Input Processing and Embedding}

To mitigate distribution shifts across distinct time series and stabilize the optimization process, the raw input sequence $\mathbf{x}$ is first standardized via standard instance normalization. Following the UPPMS, the sequence is represented by a sequence of $T$ non-overlapping patches $\mathcal{P} = \{\mathbf{p}_1, \dots, \mathbf{p}_T\}$

Each patch $\mathbf{p}_i$, whether it consists of uncorrupted historical data, stochastic perturbations, or the predictive query embedding $\mathbf{E}_{[FORECAST]}$, is linearly projected into a uniform $d$-dimensional embedding space, yielding a sequence of tokens $\mathbf{Z} \in \mathbb{R}^{T \times d}$. Temporal order is then captured by Rotary Position Embedding (RoPE) applied to the attention queries and keys within the HA Encoder.

\begin{table*}[ht]
	\centering
        \caption{Long-term forecasting task. Results are averaged over four prediction horizons: \{96, 192, 336, 720\}. Other models all adhere to the core research conclusions proposed in the original papers. The best results are represented in \textbf{bold}.}
        \label{tab_ltf}
   \setlength{\tabcolsep}{5pt}
   \renewcommand{\arraystretch}{1}
	\resizebox{\linewidth}{!}{	
		\begin{tabular}{c|cc|cc|cc|cc|cc|cc|cc|cc|cc|cc}
			\toprule[1.0pt]
			Methods & \multicolumn{2}{c|}{TimeHAT} 
            &
            \multicolumn{2}{c|}{MOMENT} &
            \multicolumn{2}{c|}{GPT4TS} & \multicolumn{2}{c|}{TimesNet} & \multicolumn{2}{c|}{PatchTST} & \multicolumn{2}{c|}{Timer-XL} & \multicolumn{2}{c|}{iTransformer} & \multicolumn{2}{c|}{CrossLinear} & \multicolumn{2}{c|}{TimeMixer} & \multicolumn{2}{c}{TimeMixer++}
            \\ 
			\midrule[0.5pt]
			Metric                       & MSE              & MAE             & MSE   & \multicolumn{1}{c|}{MAE}   & MSE           & MAE           & MSE           & MAE           & MSE           & MAE           & MSE             & MAE            & MSE   & MAE           & MSE              & MAE    & MSE              & MAE   & MSE              & MAE    \\ 
     			\midrule[0.5pt]
ETTh$_1$ & \textbf{0.402} & \textbf{0.421} & 0.418        & 0.436        & 0.427        & 0.426        & 0.458         & 0.450         & 0.413         & 0.430         & 0.409         & 0.430        & 0.454           & 0.447           & 0.431           & 0.433          & 0.447          & 0.440         & 0.419          & 0.432          \\ \midrule[0.5pt]
ETTh$_2$ & \textbf{0.322} & \textbf{0.368} & 0.352        & 0.395        & 0.346        & 0.394        & 0.414         & 0.427         & 0.330         & 0.379         & 0.352         & 0.402        & 0.383           & 0.407           & 0.368           & 0.396          & 0.364          & 0.395         & 0.339          & 0.380          \\ \midrule[0.5pt]
ETTm$_1$ & \textbf{0.336} & \textbf{0.366} & 0.344        & 0.379        & 0.352        & 0.383        & 0.400         & 0.406         & 0.351         & 0.387         & 0.359         & 0.382        & 0.407           & 0.410           & 0.370           & 0.393          & 0.369          & 0.378         & 0.369          & 0.378          \\ \midrule[0.5pt]
ETTm$_2$ & \textbf{0.252} & \textbf{0.309} & 0.259        & 0.318        & 0.266        & 0.326        & 0.291         & 0.333         & 0.255         & 0.315         & 0.271         & 0.322        & 0.288           & 0.332           & 0.272           & 0.320          & 0.275          & 0.323         & 0.269          & 0.320          \\ \midrule[0.5pt]
Weather  & \textbf{0.221} & \textbf{0.255} & 0.228        & 0.270        & 0.237        & 0.270        & 0.259         & 0.287         & 0.225         & 0.264         & 0.240         & 0.273        & 0.258           & 0.279           & 0.238           & 0.269          & 0.240          & 0.271         & 0.226          & 0.262          \\ \midrule[0.5pt]
Electricity      & \textbf{0.152} & \textbf{0.241} & 0.165        & 0.260        & 0.167        & 0.263        & 0.192         & 0.295         & 0.161         & 0.253         & 0.155         & 0.246        & 0.178           & 0.270           & 0.174           & 0.270          & 0.182          & 0.272         & 0.165          & 0.253          \\ 
			\bottomrule[1.0pt] 
		\end{tabular}
}	
\end{table*}

\subsection{Hybrid Attention Encoder}
\label{sec:hybrid_attention}

To mitigate rank degeneration while preserving effective contextual
information exchange, we develop a Hybrid Attention Encoder (HAE) with
a progressive attention architecture. The first $M$ layers employ
prefix-constrained bidirectional attention (BiA) to integrate historical
context, while the remaining $L_e-M$ layers employ directional hybrid attention
to improve rank stability in deeper layers.
Throughout the encoder, prefix constraints prevent historical tokens
from accessing predictive targets and preserve causal information flow
among target tokens.

\subsubsection{Unified Formulation and Notation.}

Let $L_e$ denote the number of encoder layers, $T$ the number of input
tokens, and $d_{\mathrm{model}}$ the hidden dimension. We use an even
number $h$ of attention heads, each with dimension
$d_k=d_{\mathrm{model}}/h$. Let
$\mathbf{X}^{(l-1)}\in\mathbb{R}^{T\times d_{\mathrm{model}}}$
denote the input to layer $l$. For each head $\alpha$, the query, key,
and value matrices are computed as
$\mathbf{Q}_{\alpha}^{(l)}=\mathbf{X}^{(l-1)}\mathbf{W}_{Q,\alpha}^{(l)}$,
$\mathbf{K}_{\alpha}^{(l)}=\mathbf{X}^{(l-1)}\mathbf{W}_{K,\alpha}^{(l)}$,
and $\mathbf{V}_{\alpha}^{(l)}=\mathbf{X}^{(l-1)}\mathbf{W}_{V,\alpha}^{(l)}$,
with projection matrices in $\mathbb{R}^{d_{\mathrm{model}}\times d_k}$.
The attention scores are
$\mathbf{S}_{\alpha}^{(l)}
=\mathbf{Q}_{\alpha}^{(l)}{\mathbf{K}_{\alpha}^{(l)}}^\top/\sqrt{d_k}$,
and the head output is
$\mathbf{H}_{\alpha}^{(l)}
=\operatorname{Softmax}(\mathbf{S}_{\alpha}^{(l)}
+\mathbf{M}_{\alpha}^{(l)})\mathbf{V}_{\alpha}^{(l)}$,
where $\mathbf{M}_{\alpha}^{(l)}\in\{0,-\infty\}^{T\times T}$
is the stage- and head-specific attention mask defined below.

\paragraph{Stage 1: Prefix-Constrained BiA Layers ($l=1,\ldots,M$).}

The first $M$ layers apply bidirectional attention among non-target tokens
while enforcing causal information flow for predictive targets. We use
the prefix constraint mask $\mathbf{M}^{\mathrm{prefix}}$ defined in
Section~\ref{sec:uppms}, setting
$\mathbf{M}_{\alpha}^{(l)}=\mathbf{M}^{\mathrm{prefix}}$ for every
attention head $\alpha$. This allows non-target tokens to interact
bidirectionally, while each target token can access historical context
and preceding target tokens, including itself, without accessing
subsequent targets.

\paragraph{Stage 2: Directional Hybrid Attention Layers
($l=M+1,\ldots,L_e$).}

In deeper layers, we introduce complementary directional attention
to mitigate rank degeneration. We evenly divide the $h$ attention
heads into forward and backward subsets,
$\mathcal{H}_{\mathrm{fw}}$ and $\mathcal{H}_{\mathrm{bw}}$,
with $|\mathcal{H}_{\mathrm{fw}}|
=|\mathcal{H}_{\mathrm{bw}}|=h/2$.
The attention masks are defined by combining directional constraints
with the prefix constraint:
\begin{equation}
\mathbf{M}_{\alpha}^{(l)} =
\begin{cases}
\mathbf{M}^{\mathrm{causal}}+\mathbf{M}^{\mathrm{prefix}},
& \alpha\in\mathcal{H}_{\mathrm{fw}},\\
\mathbf{M}^{\mathrm{anti}}+\mathbf{M}^{\mathrm{prefix}},
& \alpha\in\mathcal{H}_{\mathrm{bw}}.
\end{cases}
\end{equation}

The causal mask follows the standard lower-triangular constraint,
whereas the anti-causal mask adapts to the target positions:
\begin{equation}
M_{ij}^{\mathrm{causal}} =
\begin{cases}
0, & j\leq i,\\
-\infty, & j>i,
\end{cases}
\qquad
M_{ij}^{\mathrm{anti}} =
\begin{cases}
0, & i\notin\mathcal{I}_{\mathrm{target}},\ j\geq i,\\
0, & i\in\mathcal{I}_{\mathrm{target}},\ j\leq i,\\
-\infty, & \text{otherwise}.
\end{cases}
\end{equation}

Thus, forward heads use causal attention throughout, while backward
heads use anti-causal attention for historical queries and switch to
causal attention for predictive targets. Combined with the prefix
constraint, this design preserves access to historical context for
all target queries while preventing information leakage from future
targets. The resulting attention weight matrices are full-rank under these
masking constraints, owing to their triangular or block-triangular
structure and positive diagonal entries.

\subsubsection{Layer Integration.}

For each layer, the outputs of all $h$ attention heads are concatenated
and projected to the model dimension:
\begin{equation}
\widetilde{\mathbf{X}}^{(l)}
=
\operatorname{Concat}\big(
\mathbf{H}_{1}^{(l)},\ldots,\mathbf{H}_{h}^{(l)}
\big)\mathbf{W}_{O}^{(l)},
\end{equation}
where $\mathbf{W}_{O}^{(l)}
\in\mathbb{R}^{h d_k\times d_{\mathrm{model}}}$.
The projected output is combined with the layer input through a residual
connection, followed by the FFN and its residual connection, to obtain
the layer output $\mathbf{X}^{(l)}$.

The directional hybrid attention layers use full-rank attention
matrices to mitigate rank degeneration during deeper information
exchange. Setting $M=0$ yields an encoder composed entirely of
directional hybrid-attention layers.

\begin{table*}[ht]
	\centering 
        \caption{Imputation task. We randomly mask the input time series at \{12.5\%, 25\%, 37.5\%, 50\%\}. The results are averaged from 4 different mask ratios. The best performance is shown in \textbf{Bold}.}
        \label{tab_it}
  \setlength{\tabcolsep}{5pt}
   \renewcommand{\arraystretch}{1}
	\resizebox{\linewidth}{!}{	
		\begin{tabular}{c|cc|cc|cc|cc|cc|cc|cc|cc|cc|cc}
			\toprule[1.0pt]
			Methods & \multicolumn{2}{c|}{TimeHAT} & \multicolumn{2}{c|}{GPT4TS} & \multicolumn{2}{c|}{TimesNet} & \multicolumn{2}{c|}{PatchTST} & \multicolumn{2}{c|}{ETSformer} & \multicolumn{2}{c|}{LightTS} & \multicolumn{2}{c|}{DLinear} & \multicolumn{2}{c|}{FEDformer} & \multicolumn{2}{c|}{Non-Stationary} & \multicolumn{2}{c}{Autoformer}\\ 
			\midrule[0.5pt]
			Metric                       & MSE              & MAE             & MSE   & \multicolumn{1}{c|}{MAE}   & MSE           & MAE           & MSE           & MAE           & MSE           & MAE           & MSE             & MAE            & MSE   & MAE           & MSE              & MAE    & MSE              & MAE   & MSE              & MAE    \\ 
     			\midrule[0.5pt]
ETTm$_1$
&\textbf{0.021}&\textbf{0.095}&0.028&0.105&0.027&0.107&0.047&0.140&0.120&0.253&0.104&0.218&0.093&0.206&0.062&0.177&0.036&0.126&0.051&0.150  \\ 
   			\midrule[0.5pt]
ETTm$_2$
&\textbf{0.015}&\textbf{0.076}&0.021&0.084&0.022&0.088&0.029&0.102&0.208&0.327&0.046&0.151&0.096&0.208&0.101&0.215&0.026&0.099&0.029&0.105  \\ 
   			\midrule[0.5pt]
			ETTh$_1$    
&\textbf{0.052}&\textbf{0.151}&0.069&0.173&0.078&0.187&0.115&0.224&0.202&0.329&0.284&0.373&0.201&0.305&0.17&0.246&0.094&0.201&0.103&0.314  \\ 
   			\midrule[0.5pt]
			ETTh$_2$
&\textbf{0.041}&\textbf{0.134}&0.048&0.141&0.049&0.146&0.065&0.163&0.367&0.436&0.119&0.250&0.142&0.259&0.163&0.279&0.053&0.152&0.055&0.156   \\ 
   	\midrule[0.5pt]
			Electricity 
&\textbf{0.066}&\textbf{0.182}&0.090&0.207&0.092&0.210&0.072&0.183&0.214&0.339&0.131&0.262&0.132&0.260&0.130&0.259&0.100&0.218&0.101&0.225 \\
   	\midrule[0.5pt]
			Weather                      
&\textbf{0.026}&\textbf{0.047}&0.031&0.056&0.030&0.054&0.034&0.055&0.076&0.171&0.055&0.117&0.052&0.110&0.099&0.203&0.032&0.059&0.031&0.057 \\	
			\bottomrule[1.0pt] 
		\end{tabular}
}	
\end{table*}

\section{Experiments}

We pre-train TimeHAT on UTSD-12G \cite{liu2024timer}, a comprehensive corpus comprising 1 billion time points across diverse domains. For downstream evaluations, spanning forecasting, imputation, classification, and anomaly detection, we strictly follow the standardized protocols established by TimesNet \cite{wu2022timesnet}.

\textbf{Baselines:} 
The baselines include CNN-based models: TimesNet \cite{wu2022timesnet}, TimeMixer++\cite{wang2025timemixer++}; 
MLP-based models: LightTS \cite{zhang2022less}, DLinear \cite{zeng2023transformers}, TimeMixer\cite{wangtimemixer}; 
Transformer-based models: Timer-XL \cite{liutimer}, MOMENT \cite{goswamimoment}, GPT4TS\cite{zhou2023one}, iTransformer \cite{liuitransformer}, Autoformer \cite{wu2021autoformer}, FEDformer \cite{zhou2022fedformer}, Non-stationary Transformer \cite{liu2022non}, ETSformer \cite{woo2022etsformer}, PatchTST \cite{nie2022time}, TIMESBERT \cite{zhang2025timesbert}, Anomaly Transformer \cite{xu2021anomaly}, Pyraformer \cite{liu2021pyraformer}. 
Besides, Rocket \cite{dempster2020rocket}, LSTNet \cite{lai2018modeling}, LSSL \cite{gu2021efficiently}, TCN \cite{franceschi2019unsupervised} and Flowformer \cite{huang2022flowformer} are used for classification.

\subsection{Long-term Forecasting}

We evaluate long-term forecasting across six benchmarks, reporting MSE and MAE averaged over $\{96, 192, 336, 720\}$.
As detailed in Table \ref{tab_ltf}, TimeHAT achieves competitive performance, achieves the best average performance on the reported benchmarks. Notably, it demonstrates a decisive advantage over both highly optimized forecasting architectures (e.g., TimeMixer++, iTransformer) and recent large-scale time series foundation models (e.g., MOMENT, GPT4TS). This superiority stems from three synergistic mechanisms: (1) Reconstructive Sampling masks the terminal token to enforce robust next-step prediction from noisy histories; (2) Extrapolative Sampling masks continuous future sequences, optimizing the global bottleneck for extended horizons; and (3) the Hybrid Attention Encoder structurally prevents depth-induced rank collapse. Together, they seamlessly translate full-rank contextual features into highly accurate long-term extrapolations.

\subsection{Imputation}

We evaluate the models using a fixed input length of 96, applying random masking at four distinct ratios: $\{12.5\%, 25\%, 37.5\%, 50\%\}$. 
As reported in Table \ref{tab_it}, TimeHAT achieves the lowest averaged MSE and MAE across the six benchmarks, substantially outperforming several strong
baselines such as GPT4TS. This reconstructive superiority validates our Reconstructive Sampling and prefix-constrained BiA layers. Preserving full-rank, unconstrained interactions within the observable context allows TimeHAT to leverage global contextual competition for precise interpolation of missing segments.

\begin{table}[ht]
\centering
        \caption{Anomaly detection task. We calculate the F1-score (as \%) for each dataset. 
\textbf{Bold} represents the best.}
        \label{tab_adt}
  \setlength{\tabcolsep}{5pt}
   \renewcommand{\arraystretch}{1}
	\resizebox{0.9\linewidth}{!}{	
\begin{tabular}{c|ccccc|c}
	\toprule[1.0pt]
Methods           & SMD          & MSL                     & SMAP                    & SWaT                    & PSM                     &  Average \\\midrule[0.5pt]
TimeHAT & 86.95  & 89.46  & 76.36   & 94.64    & 97.71 & \textbf{89.02}    \\ \midrule[0.5pt]
TIMESBERT                          & 86.04               & 88.07                   & 75.69                   & 93.95                   & 98.27                   & 88.40               \\ \midrule[0.5pt]
GPT4TS                          & 86.89                & 82.45                   & 72.88                   & 94.23                   & 97.13                   & 86.72               \\ \midrule[0.5pt]
TimesNet                     & 84.61                   & 81.84                   & 69.39                   & 93.02                   & 97.34                   & 85.24               \\ \midrule[0.5pt]
Pyraformer                   & 83.04                   & 84.86                   & 71.09                   & 91.78                   & 82.08                   & 82.57               \\ \midrule[0.5pt]
\begin{tabular}[c]{@{}c@{}}Anomaly\\ Transformer\end{tabular}                     & 85.49                   & 83.31                   & 71.18                   & 83.10                   & 79.40                   & 80.50               \\ \midrule[0.5pt]
Informer                     & 81.65                   & 84.06                   & 69.92                   & 81.43                   & 77.10                   & 78.83               \\ 
\midrule[0.5pt]
Reformer                      & 75.32                   & 84.40                   & 70.40                   & 82.80                   & 73.61                   & 77.31               \\ 
\midrule[0.5pt]
LogTrans                      & 76.21                   & 79.57                   & 69.97                   & 80.52                   & 76.74                   & 76.60               \\ 
\midrule[0.5pt]
Transformer                      & 79.56                   & 78.68                   & 69.70                   & 80.37                   & 76.07                   & 76.88               \\ 
\bottomrule[1.0pt] 
\end{tabular}
}
\end{table}

\subsection{Anomaly Detection}

We evaluate the capability of TimeHAT in identifying abnormal temporal patterns across five standard benchmarks: SMD, MSL, SMAP, SWaT, and PSM. Following standard protocols, the F1-score is adopted as the primary evaluation metric. 
As summarized in Table \ref{tab_adt}, TimeHAT consistently outperforms specialized architectures (e.g., Anomaly Transformer) and recent foundation models (e.g., TIMESBERT). This superiority stems directly from our pre-training paradigm: by simultaneously modeling high-frequency micro-variations and macroscopic trends, the global representation bottleneck exhibits exceptional sensitivity to both structural deviations and localized anomalies.

\subsection{Classification}

\begin{figure}[h]
 \centering
	\includegraphics[scale=0.35]{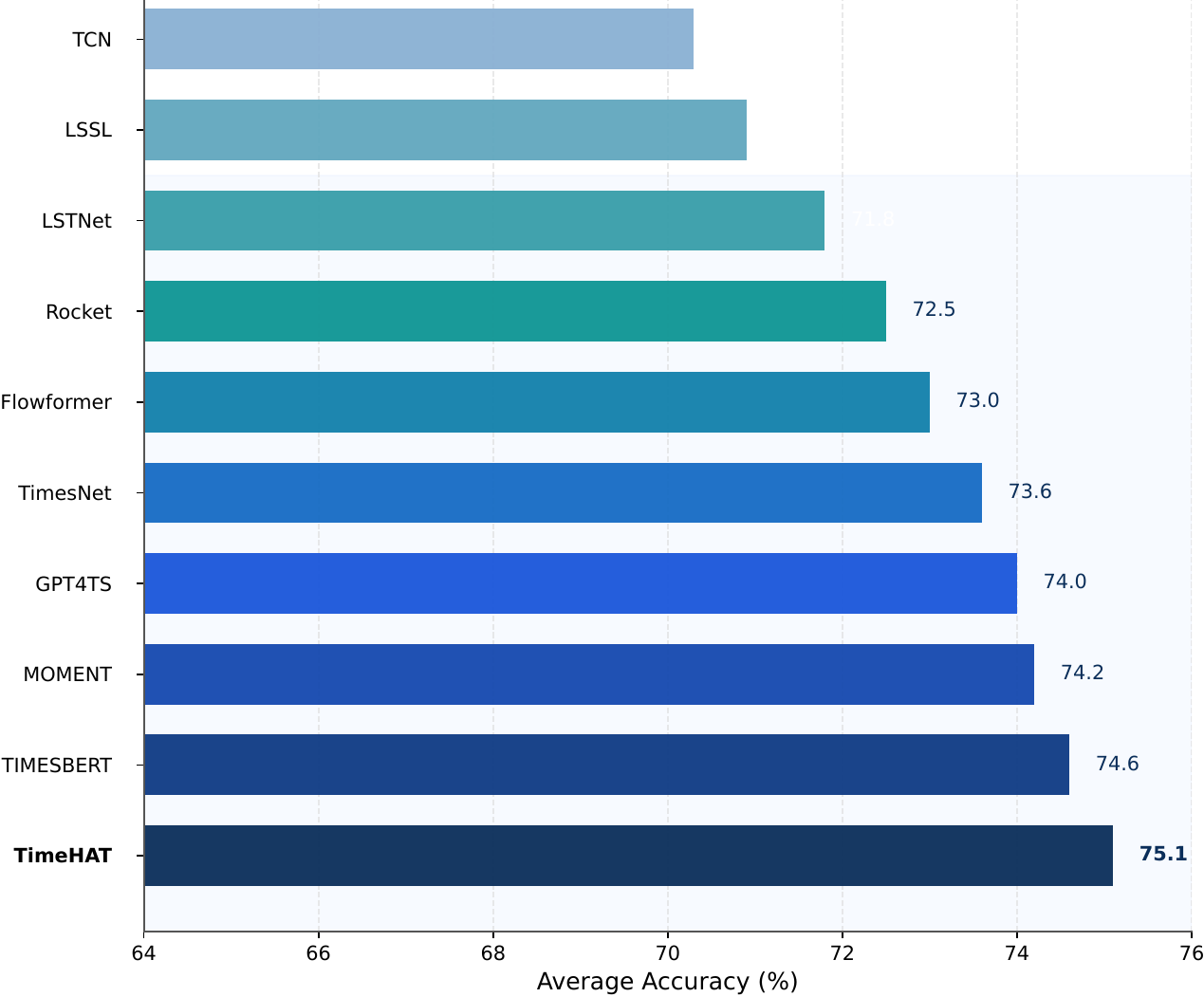}
  \caption{Model comparison in classification.}
  \label{fig_c}
\end{figure}

To adapt the pre-trained TimeHAT for time series classification, we naturally leverage the terminal token, previously established as the global representation bottleneck, to serve as a \texttt{[CLS]}-equivalent semantic probe. For multivariate inputs, an additional MLP head is appended to the terminal representation, effectively fusing cross-variate dependencies and projecting the unified features into the target class space.
We comprehensively fine-tune the pre-trained model on 10 benchmark datasets from the UEA Multivariate Time Series Classification Archive, which encompass a diverse range of application scenarios. Utilizing classification accuracy as the primary evaluation metric, empirical results validate the superior generalization capability of our architecture. As illustrated in Figure \ref{fig_c}, TimeHAT achieves the highest average classification accuracy across all 10 evaluated datasets.

\subsection{Zero-shot and Few-shot Forecasting}

\begin{table}[htbp]
\centering
\caption{Zero-shot capability (MSE).}
\renewcommand{\arraystretch}{1}
	\resizebox{0.9\linewidth}{!}{	
\begin{tabular}{c|c|c|c|c|c}
\toprule[1.0pt]
Dataset & TimeHAT & GTM & Timer & Moirai & MOMENT \\
\midrule[0.5pt]
ETTh1 & \textbf{0.399} & 0.407 & 0.438 & 0.441 & 0.674 \\
ETTm1 & \textbf{0.558} & 0.593 & 0.690 & 0.562 & 0.670 \\
Weather & \textbf{0.170} & 0.172 & 0.181 & 0.195 & 0.255 \\
Electricity & \textbf{0.176} & 0.187 & 0.192 & 0.212 & 0.744 \\
Traffic & \textbf{0.447} & 0.542 & 0.458 & 0.616 & 1.293 \\
\bottomrule[1.0pt] 
\end{tabular}
}
\label{tab:perf_compare}
\end{table}

\begin{figure}[htbp] 
  \centering
  
  \begin{subfigure}[b]{0.48\linewidth}
    \centering
    \includegraphics[width=\linewidth]{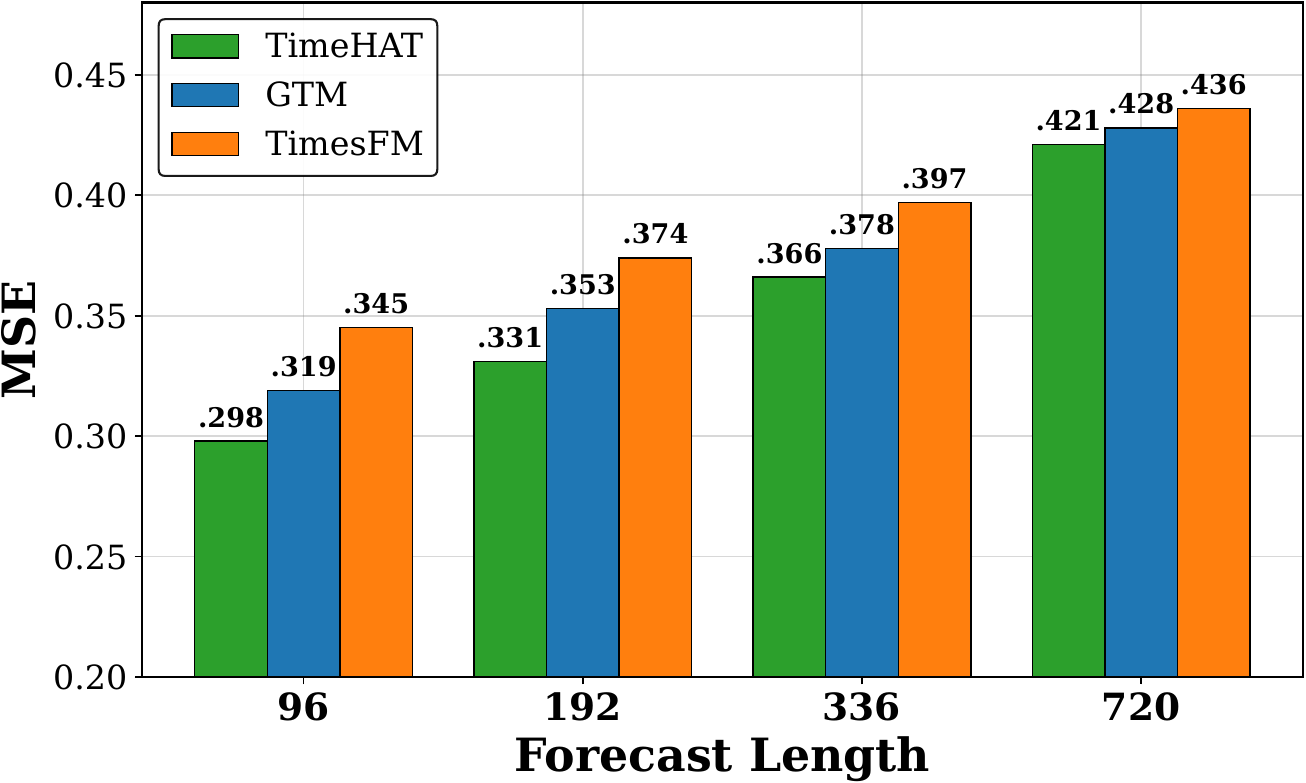}
  \end{subfigure}
  \begin{subfigure}[b]{0.48\linewidth}
    \centering
    \includegraphics[width=\linewidth]{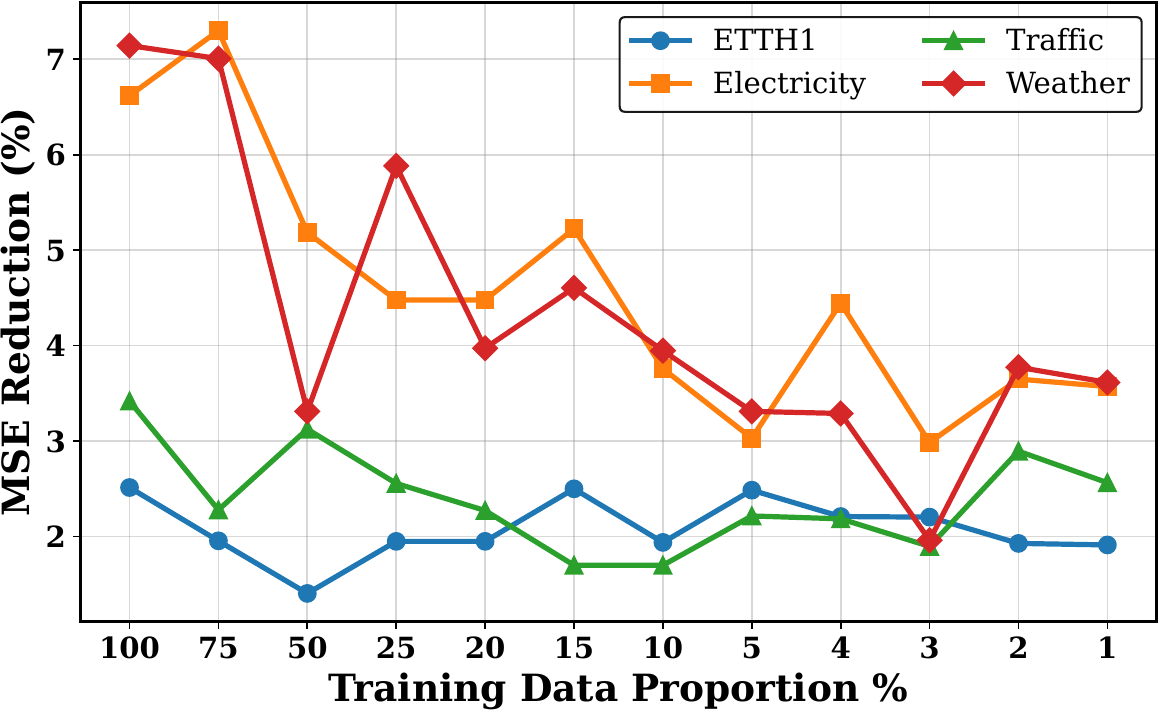}
  \end{subfigure}

  \caption{Few-shot capability of TimeHAT compared to baselines.}
  \label{fig_overall_evaluation}
\end{figure}

To evaluate generalization and data efficiency, Table~\ref{tab:perf_compare} details TimeHAT's zero-shot performance against GTM \cite{he2025general}, Timer \cite{liu2024timer}, Moirai \cite{woo2024unified}, and MOMENT \cite{goswamimoment}. Figure~\ref{fig_overall_evaluation} illustrates its few-shot capabilities: the left compares MSE across varying forecast lengths against GTM and TimesFM \cite{Das2024decoder}, while the right demonstrates the performance improvement over Timer as training data decreases. Overall, TimeHAT consistently outperforms all baselines, proving its robust generalization in data-scarce settings.

\subsection{Pretraining Methods Analysis}

\subsubsection{Direct versus Recursive Forecasting}

\begin{figure}[htbp]
    \centering
    \includegraphics[scale=0.22]{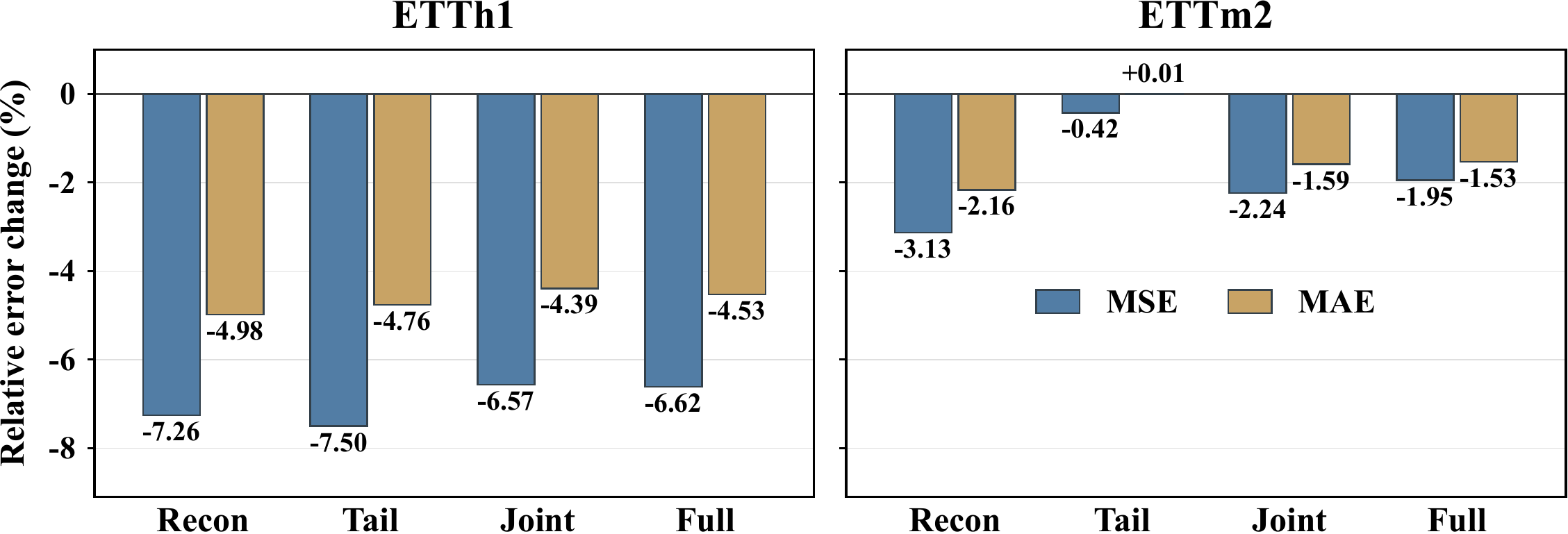}
    \caption{Relative error changes of recursive versus direct forecasting
across pre-training schemes on ETTh1 and ETTm2.
Recon: Contextual Reconstruction; Tail: Causal Extrapolation;
Joint: joint objectives; Full: full UPPMS.}
\label{fig:direct_vs_recursive}
\end{figure}

We compare direct forecasting, which predicts the entire target horizon
in one pass, with recursive forecasting, which repeatedly predicts
96 steps and feeds the predictions back as input to extend the forecast.
Figure~\ref{fig:direct_vs_recursive} shows that recursive forecasting
reduces MSE across all evaluated schemes on both datasets. 
On ETTh1, MSE decreases by 6.57\%--7.50\%, with MAE reductions of 4.39\%--4.98\%.
These results support recursive forecasting for downstream evaluation.
It also allows the pre-trained reconstruction head to be reused,
preserving consistency between pre-training and downstream prediction
without reinitializing the head for different forecasting horizons.

\begin{table}[h]
    \centering
    \caption{Relative changes in forecasting errors under UPPMS ablations
(\%). Positive values indicate higher errors than the full method.
LPR: w/o last-patch recovery; LPA: w/o left-padding augmentation;
PO: patch-only masking; ZO: zero-only replacement;
FTL: fixed tail length.}
    \label{tab:component_ablation}
    \resizebox{\linewidth}{!}{
    \begin{tabular}{c|rr|rr|rr|rr}
        \toprule
        \multirow{2}{*}{Variant}
        & \multicolumn{2}{c|}{ETTh2}
        & \multicolumn{2}{c|}{ETTm1}
        & \multicolumn{2}{c|}{ETTm2}
        & \multicolumn{2}{c}{Weather} \\
        & MSE & MAE & MSE & MAE & MSE & MAE & MSE & MAE \\
        \midrule
        LPR
        & +2.05 & +0.79 & -0.39 & +0.01
        & +0.80 & -0.16 & -0.31 & -0.25 \\
        LPA
        & +1.55 & +0.45 & +0.02 & -0.09
        & -0.45 & -0.79 & +0.17 & -0.08 \\
        PO
        & +1.79 & +0.31 & -0.37 & -0.20
        & +0.40 & -0.26 & -0.70 & -0.88 \\
        ZO
        & -1.00 & -0.46 & +0.36 & +0.27
        & +1.06 & +0.27 & +0.10 & +0.07 \\
        FTL
        & -0.92 & -0.61 & +0.40 & +0.07
        & +1.52 & -1.08 & -0.07 & -0.22 \\
        \bottomrule
    \end{tabular}
    }
\end{table}

\subsubsection{Sensitivity to UPPMS Design Choices}

We compare the full UPPMS with variants that remove or simplify
individual masking mechanisms on four datasets. Table~\ref{tab:component_ablation}
reports relative changes in MSE and MAE averaged over three fine-tuning
seeds, with positive values indicating higher errors after ablation.
The results reveal dataset-dependent sensitivity to masking design.
Removing last-patch recovery or restricting masking to patches increases
MSE on ETTh2, while zero-only replacement degrades both metrics on three
of the four datasets. Other simplifications yield mixed effects across
datasets and metrics. Overall, no single simplification consistently
improves both metrics across all datasets, supporting the use of
complementary masking mechanisms in UPPMS to achieve balanced performance
across diverse time series.

\subsubsection{Effect of Pre-training Objectives}
We compare four pre-training schemes under the same six-layer encoder
and fine-tuning protocol, using the final historical token to forecast
96 future steps. Each scheme is pre-trained on the training split, and
results are averaged over three fine-tuning seeds.
As shown in Table~\ref{tab:pretraining_last_token}, the full method
achieves the lowest MSE on all five datasets and the lowest MAE on
four. The improvements are particularly notable on ETTh2, while
ETTm1 shows a slight MAE advantage for Causal Extrapolation. These results demonstrate the benefits of the full UPPMS for downstream forecasting.

\begin{table}[h]
    \centering
    \caption{Forecasting performance of different pre-training objectives.
    Full: full UPPMS; Tail: Causal Extrapolation; Joint: joint objectives;
    Random: Contextual Reconstruction.}
    \label{tab:pretraining_last_token}
    \resizebox{0.96\linewidth}{!}{
    \begin{tabular}{l|cc|cc|cc|cc}
        \toprule
        \multirow{2}{*}{Dataset}
        & \multicolumn{2}{c|}{Full}
        & \multicolumn{2}{c|}{Tail}
        & \multicolumn{2}{c|}{Joint}
        & \multicolumn{2}{c}{Random} \\
        & MSE & MAE & MSE & MAE & MSE & MAE & MSE & MAE \\
        \midrule
        ETTh1
        & \textbf{0.384} & \textbf{0.409}
        & 0.390 & 0.413
        & 0.386 & 0.410
        & 0.387 & 0.410 \\
        ETTh2
        & \textbf{0.300} & \textbf{0.353}
        & 0.321 & 0.364
        & 0.307 & 0.356
        & 0.308 & 0.356 \\
        ETTm1
        & \textbf{0.294} & 0.355
        & 0.297 & \textbf{0.353}
        & 0.301 & 0.360
        & 0.298 & 0.357 \\
        ETTm2
        & \textbf{0.177} & \textbf{0.266}
        & 0.179 & 0.269
        & 0.178 & 0.268
        & 0.177 & 0.268 \\
        Weather
        & \textbf{0.152} & \textbf{0.202}
        & 0.153 & 0.203
        & 0.154 & 0.204
        & 0.153 & 0.203 \\
        \bottomrule
    \end{tabular}
    }
\end{table}

\subsubsection{Effect of Pre-training Mode Balance}

We vary the probability of assigning a pre-training sample to tail
recovery, $p_{\mathrm{tail}}\in\{0,0.25,0.40,0.50,0.75,1\}$, with the
remaining samples assigned to random reconstruction. Here,
$p_{\mathrm{tail}}=0$ corresponds to pure random reconstruction, whereas
$p_{\mathrm{tail}}=1$ corresponds to pure tail recovery. Models are
pre-trained and fine-tuned on the same dataset to evaluate how the
balance between the two pre-training modes affects downstream forecasting.
Table~\ref{tab:tail_probability} shows that
intermediate probabilities generally yield better results than pure
tail recovery. 
Overall, combining random reconstruction and tail recovery provides
a more effective balance than relying exclusively on tail recovery.

\begin{table}[h]
    \centering
    \caption{Forecasting performance under different tail-task probabilities.}
    \label{tab:tail_probability}
    \small
    \begin{tabular}{cccccccc}
        \toprule
        Dataset & Metric
        & $0$ & $0.25$ & $0.40$
        & $0.50$ & $0.75$ & $1$ \\
        \midrule
        \multirow{2}{*}{ETTh1}
        & MSE & 0.403 & 0.394 & 0.396 & 0.395 & 0.396 & 0.402 \\
        & MAE & 0.412 & 0.411 & 0.412 & 0.411 & 0.412 & 0.416 \\
        \midrule
        \multirow{2}{*}{ETTm2}
        & MSE & 0.176 & 0.176 & 0.173 & 0.173 & 0.178 & 0.177 \\
        & MAE & 0.257 & 0.258 & 0.258 & 0.257 & 0.258 & 0.263 \\
        \midrule
        \multirow{2}{*}{Weather}
        & MSE & 0.150 & 0.150 & 0.149 & 0.149 & 0.149 & 0.152 \\
        & MAE & 0.195 & 0.195 & 0.193 & 0.193 & 0.192 & 0.195 \\
        \bottomrule
    \end{tabular}%
  
\end{table}

\subsection{Hybrid Attention Encoder Analysis}

\subsubsection{Attention Performance Across Forecast Horizons and Depth}

\begin{figure}[htbp]
    \centering
    \includegraphics[scale=0.26]{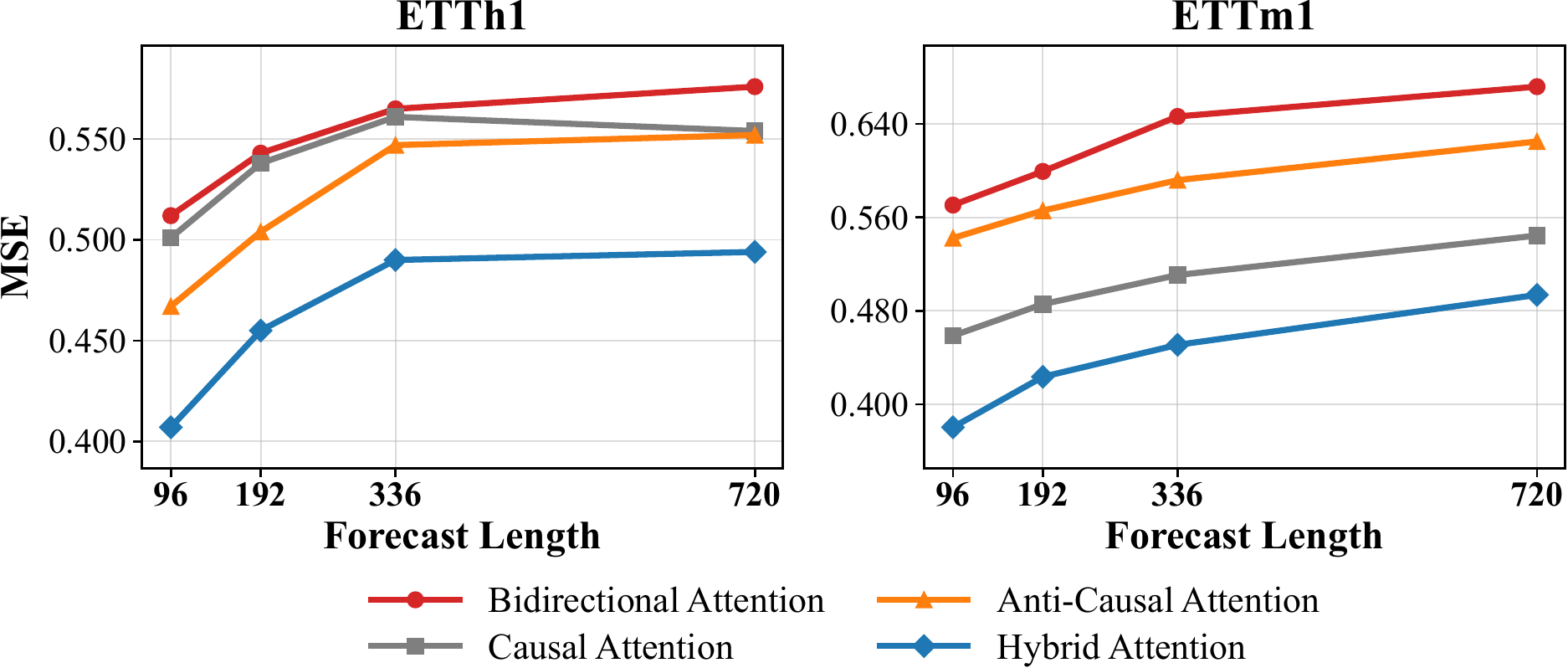}
    \caption{Forecasting MSE of different attention mechanisms across
prediction horizons on ETTh1 and ETTm1.}
\label{fig:attention_horizon}
\end{figure}

\begin{figure}[htbp]
    \centering
    \includegraphics[scale=0.26]{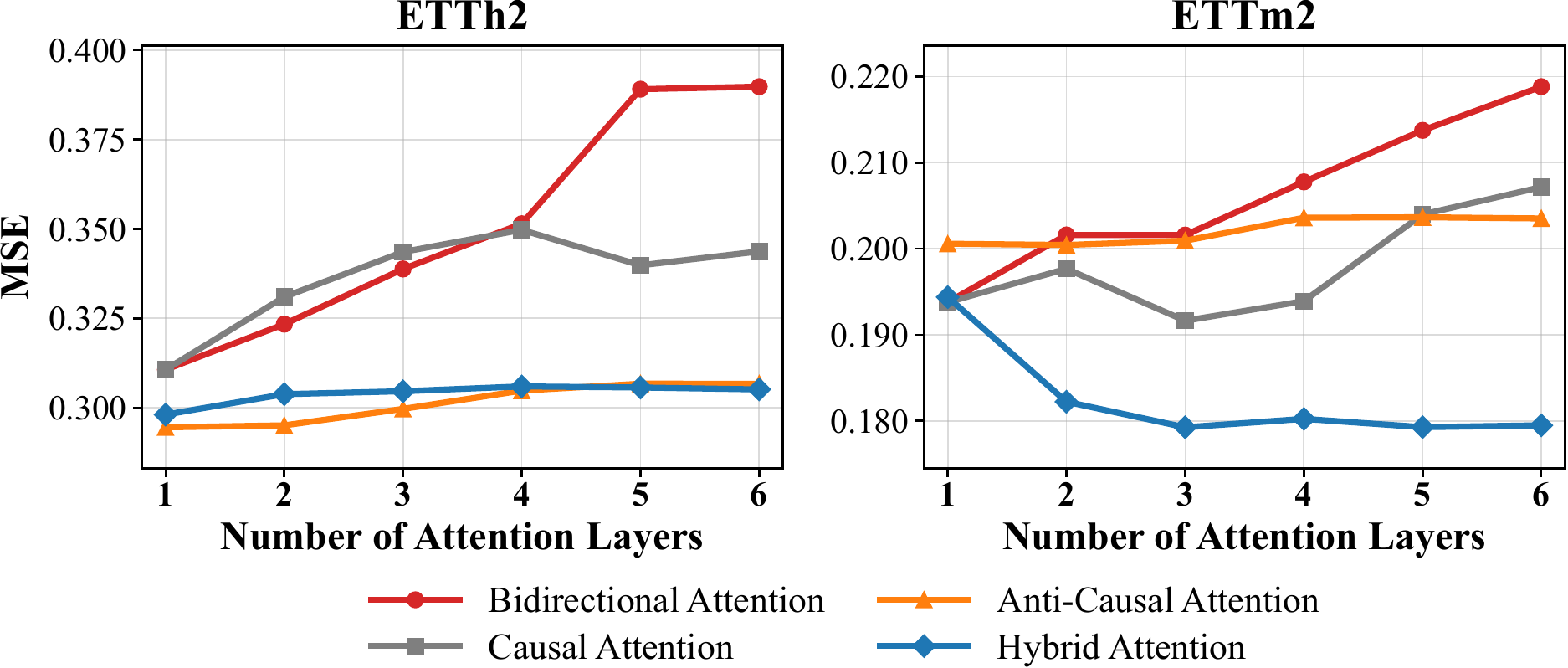}
    \caption{Forecasting MSE of different attention mechanisms across
network depths on ETTh2 and ETTm2.}
\label{fig:attention_depth}
\end{figure}

We compare BiA, causal attention, anti-causal attention, and HA across
forecast horizons and network depths. Figure~\ref{fig:attention_horizon}
shows that HA achieves the lowest MSE across the evaluated horizons on
ETTh1 and ETTm1. On ETTm1, it reduces MSE by 26.5\%--33.3\% relative
to BiA, demonstrating its strong forecasting performance across different
prediction lengths.
Figure~\ref{fig:attention_depth} examines the effect of increasing
attention depth from one to six layers on ETTh2 and ETTm2. BiA exhibits
substantial performance degradation as depth increases, whereas HA
maintains more stable performance and achieves the lowest MSE with six
layers on both datasets. Causal and anti-causal attention also exhibit greater depth robustness
than BiA, suggesting that directional masking helps mitigate
depth-related performance degradation. Notably, HA further improves
as depth increases on ETTm2, while maintaining stable performance
on ETTh2.
Overall, HA achieves strong forecasting performance across prediction
horizons and exhibits greater robustness to increasing attention depth
than the other attention mechanisms.

\subsubsection{Effect of Attention Composition}

\begin{figure}[htbp]
    \centering
    \includegraphics[scale=0.26]{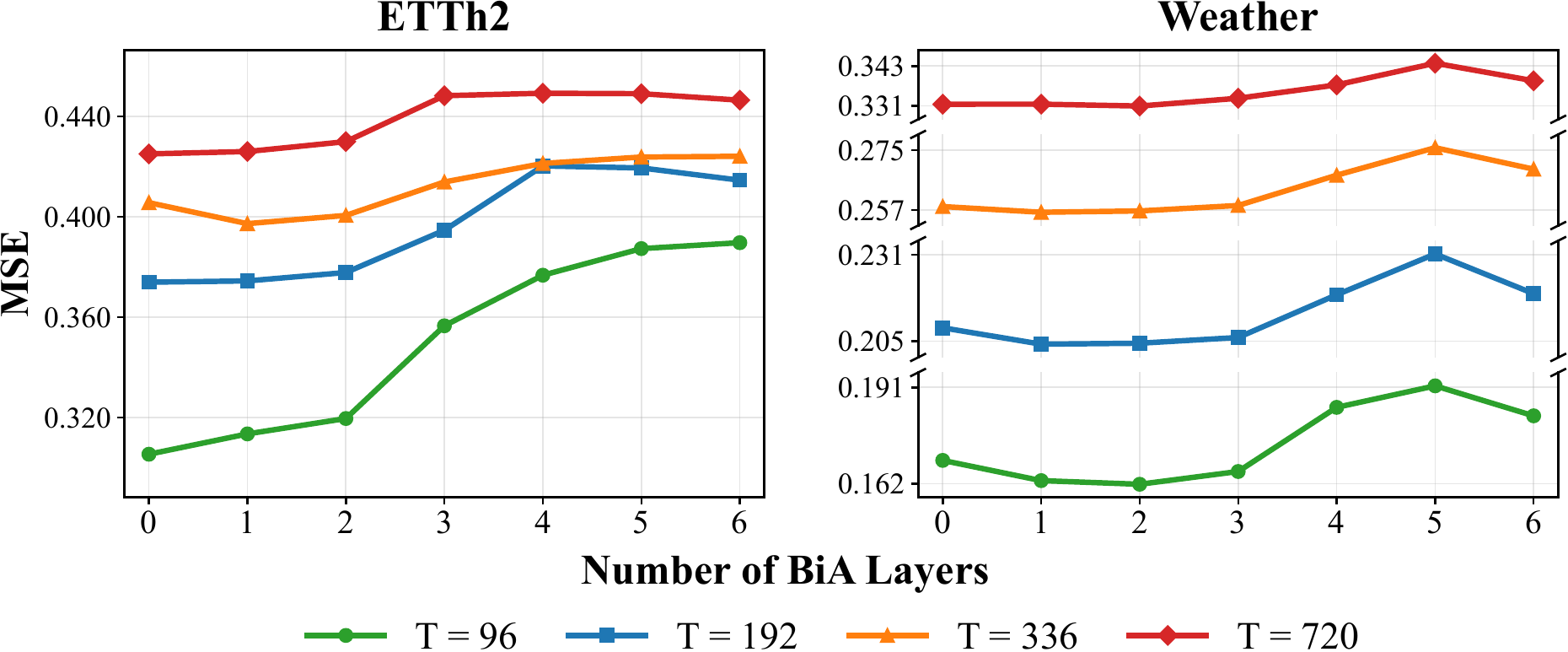}
    \caption{Forecasting MSE with different numbers of prefix-constrained
    BiA layers on ETTh2 and Weather.}
    \label{fig:prefix_bia_layers}
\end{figure}

We investigate how the number and arrangement of prefix-constrained BiA
and Mix layers affect forecasting performance. Figure~\ref{fig:prefix_bia_layers}
shows that ETTh2 maintains relatively stable performance with up to two
BiA layers at longer horizons, while Weather benefits from a small number
of BiA layers across several horizons. Increasing the BiA proportion
beyond this range can lead to performance degradation.
Table~\ref{tab:pure_attention_layer_order} shows that the all-Mix
configuration achieves the best results on the three ETT datasets, while
mixed configurations yield slightly better results on Weather. These
results demonstrate the effectiveness of Mix and show that incorporating
a limited number of BiA layers can retain competitive forecasting
performance, with additional benefits on some datasets.

\begin{table}[t]
    \centering
    \caption{Forecasting performance of different BiA--HA layer
    configurations. B and M denote BiA and HA, respectively.}
    \label{tab:pure_attention_layer_order}
    \resizebox{\linewidth}{!}{%
    \begin{tabular}{c|cc|cc|cc|cc}
        \toprule
        \multirow{2}{*}{Config}
        & \multicolumn{2}{c|}{ETTh1}
        & \multicolumn{2}{c|}{ETTm1}
        & \multicolumn{2}{c|}{ETTm2}
        & \multicolumn{2}{c}{Weather} \\
        & MSE & MAE & MSE & MAE & MSE & MAE & MSE & MAE \\
        \midrule
        MMMMMM
        & \textbf{0.507} & \textbf{0.492}
        & \textbf{0.437} & \textbf{0.440}
        & \textbf{0.278} & \textbf{0.337}
        & 0.242 & 0.280 \\
        BBBMMM
        & 0.598 & 0.541
        & 0.486 & 0.464
        & 0.287 & 0.345
        & \textbf{0.241} & 0.280 \\
        BBBBBB
        & 0.721 & 0.596
        & 0.622 & 0.527
        & 0.310 & 0.364
        & 0.252 & 0.292 \\
        MMMBBB
        & 0.538 & 0.508
        & 0.444 & 0.444
        & 0.280 & 0.339
        & 0.249 & 0.285 \\
        BMBMBM
        & 0.683 & 0.574
        & 0.442 & 0.443
        & 0.286 & 0.343
        & 0.241 & \textbf{0.280} \\
        MBMBMB
        & 0.663 & 0.562
        & 0.450 & 0.447
        & 0.287 & 0.346
        & 0.252 & 0.288 \\
        \bottomrule
    \end{tabular}%
    }
\end{table}

\subsubsection{Sensitivity to Readout Strategy}

\begin{figure}[h]
    \centering
    \includegraphics[scale=0.21]{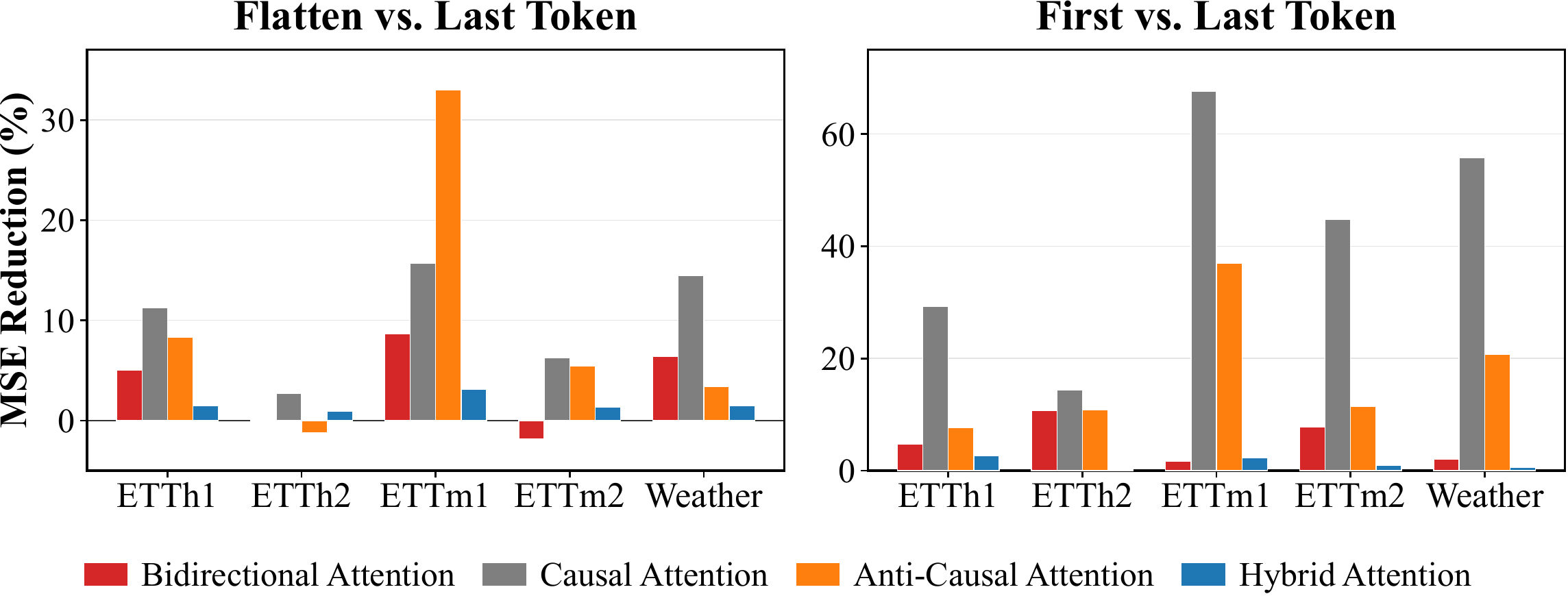}
    \caption{Sensitivity to readout strategies across attention mechanisms.}
\label{fig:readout_sensitivity}
\end{figure}

We compare different attention mechanisms under two readout settings.
We compare Flatten-Linear and Last-Token
prediction across four forecasting horizons, and First-Token and Last-Token readouts at a prediction horizon of 96.
Figure~\ref{fig:readout_sensitivity} shows that HA exhibits
small performance differences across readout strategies in both settings.
In particular, Last-Token prediction achieves results close to
Flatten-Linear, indicating that the final token captures much of the
predictive information available across the sequence. HA Attention
also shows the smallest First- versus Last-Token MSE gap in the standard
encoder. These results suggest that HA provides effective
contextual representations with relatively low sensitivity to the choice
of readout position.

\subsubsection{Effect of the Causal-Head Ratio in Hybrid Attention}

\begin{figure}[h]
 \centering
	\includegraphics[scale=0.32]{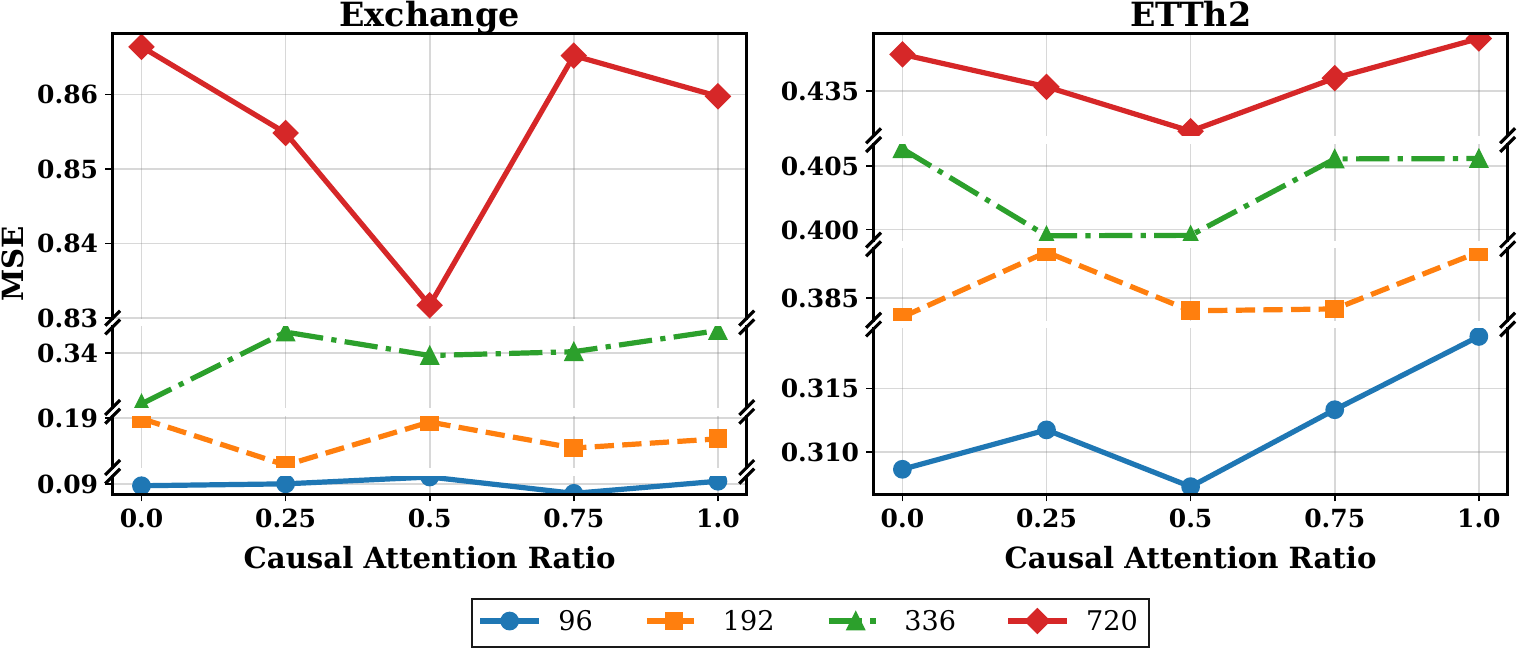}
  \caption{Performance of Different Proportions of Causal Attention in Hybrid Attention.}
  \label{fig_forward_ratio_comparison_2datasets}
\end{figure}

To investigate the impact of the CA proportion in HA, we adjusted the CA ratio $\alpha_f \in \{0, 0.25, 0.5, 0.75, 1\}$, while keeping all other hyperparameters unchanged. Here, $\alpha_f = 0$ and $1$ indicate that all heads adopt Anti-Causal and CA, respectively.
As shown in Figure \ref{fig_forward_ratio_comparison_2datasets}: when $\alpha_f = 0.5$, the model achieves the best performance on the Exchange and ETTh2 datasets, indicating that a balanced bidirectional information flow contributes to improved forecasting accuracy. 

\subsubsection{Effect of Pre-training Attention Rules}

We evaluate the impact of pre-training attention rules on downstream
forecasting by comparing the default configuration with variants that
remove task-specific constraints or modify information flow.
Table~\ref{tab:pretraining_attention_rules} shows that the default rules
achieve the lowest MSE across all four ETT datasets and the lowest MAE
on three. Removing task-specific rules or disabling direction switching
consistently increases forecasting errors. Allowing historical queries
to attend to forecast tokens also increases MSE across all four datasets,
although it slightly improves MAE on ETTm2.
These results highlight the importance of task-specific attention
constraints in pre-training, particularly for maintaining appropriate
information flow and improving downstream forecasting performance.

\begin{table}[h]
    \centering
    \caption{Downstream forecasting performance under different
    pre-training attention rules. TSR: task-specific rules;
    Hist$\rightarrow$F: historical queries can attend to forecast tokens;
    DS: direction switching for forecast queries.}
    \label{tab:pretraining_attention_rules}
    \resizebox{\linewidth}{!}{%
    \begin{tabular}{c|cc|cc|cc|cc}
        \toprule
        \multirow{2}{*}{Attention Rule}
        & \multicolumn{2}{c|}{ETTh1}
        & \multicolumn{2}{c|}{ETTh2}
        & \multicolumn{2}{c|}{ETTm1}
        & \multicolumn{2}{c}{ETTm2} \\
        & MSE & MAE & MSE & MAE & MSE & MAE & MSE & MAE \\
        \midrule
        Default
        & \textbf{0.386} & \textbf{0.410}
        & \textbf{0.307} & \textbf{0.355}
        & \textbf{0.298} & \textbf{0.357}
        & \textbf{0.177} & 0.268 \\
        w/o TSR
        & 0.389 & 0.413
        & 0.310 & 0.359
        & 0.304 & 0.362
        & 0.183 & 0.267 \\
        Hist$\rightarrow$F
        & 0.387 & 0.410
        & 0.308 & 0.356
        & 0.301 & 0.359
        & 0.179 & \textbf{0.267} \\
        w/o DS
        & 0.388 & 0.413
        & 0.310 & 0.359
        & 0.300 & 0.359
        & 0.179 & 0.267 \\
        \bottomrule
    \end{tabular}%
    }
\end{table}

\section{Conclusion}
We propose TimeHAT, a general-purpose pre-training framework for time-series analysis. Its Unified Point-Patch Masking Strategy (UPPMS) integrates point-level and patch-level supervision through contextual reconstruction and causal extrapolation, enabling the model to learn complementary temporal patterns at different granularities. Its Hybrid Attention Encoder (HAE) combines prefix-constrained bidirectional attention in initial layers with directional hybrid attention in deeper layers, facilitating contextual information exchange while mitigating rank degeneration. Extensive experiments across classification, long-term forecasting, imputation, and anomaly detection demonstrate competitive performance in fine-tuning, few-shot, and zero-shot transfer settings.

\clearpage

\bibliographystyle{ACM-Reference-Format}
\bibliography{sample-base}

\appendix

\section{Related Work}
\textbf{Large Language Models in Time Series:} Recent advances have explored applying large models like Large Language Models (LLMs) to TSA due to their ability to capture long-term dependencies and handle high-dimensional data \cite{zhao2023survey,awais2025foundation}. Frameworks such as PromptCast \cite{xue2023promptcast}, GPT4TS \cite{zhou2023one}, TIME-LLM \cite{jin2024time}, and LLM4TS \cite{chang2025llm4ts} adapt language models for time series forecasting by leveraging numerical-to-text transformations, patching mechanisms, and embedding techniques. TEST \cite{sun2023test} enhances time series tokenization and embedding through contrastive learning, while UniTime \cite{liu2024unitime} leverages domain instructions, modality alignment, and masking strategies to achieve efficient generalization and transfer in cross-domain time series forecasting. Despite their promising results, most of these models focus primarily on forecasting tasks, while broader applications of LLMs in TSA, such as anomaly detection and classification, remain underexplored.

\textbf{Pre-trining Models in Time Series:} Beyond LLMs, time series-specific pre-training frameworks have gained attention for their potential in learning foundational representations \cite{jin2023large}. 
This trend is exemplified by foundation models released alongside large-scale datasets, such as MOMENT\cite{goswamimoment},
Timer\cite{liu2024timer}, TIME-MOE \cite{shitime}, 
and Sundial \cite{liu2025sundial},
with subsequent works expanding along dimensions like scale (Moirai \cite{woo2024unified, liu2025moirai}), and task unification (UNITS \cite{gao2024units}).
Other methods, such as TimeMixer++ \cite{wang2025timemixer++}, MTSMAE \cite{tang2022mtsmae}, SimMTM \cite{dong2023simmtm}, PatchTST \cite{nie2022time}, and TSMixer \cite{ekambaram2023tsmixer}, focus on tasks like forecasting, reconstruction, and capturing spatio-temporal patterns through advanced techniques like masking, patch embedding, and lightweight architectures. These approaches demonstrate strong potential in TSA, addressing challenges such as memory efficiency, long-term forecasting, and inter-channel correlations.

\section{Pre-Training and Downstream Training Settings}
\subsection{Dataset}

Timer has curated the Unified Time Series Dataset (UTSD) \cite{liu2024timer}, which encompasses seven domains and includes up to 1 billion time points, organized in a hierarchical four-volume structure to facilitate research and pre-training of large models in the time series field. The Unified Time Series Dataset (UTSD) is meticulously compiled from a combination of publicly accessible online data repositories and empirical data derived from real-world machine operations. All datasets are classified into seven distinct domains by their sources: Energy, Environment, Health, Internet of Things (IoT), Nature, Transportation, and Web, with diverse sampling frequencies. UTSD is structured with hierarchical capacities, namely UTSD-1G, UTSD-2G, UTSD-4G, and UTSD-12G, where each smaller dataset is a subset of the larger ones. A larger subset means greater data difficulty and diversity. For our pre-training tasks, we directly utilized data from the UTSD-12G.

\begin{table}[ht]
\centering
\caption{Long-term forecasting dataset descriptions. The Dataset Splitting is organized in (Train,Validation,Test).}
\label{tab_ds1}
   \renewcommand{\arraystretch}{1.2}
	\resizebox{1\linewidth}{!}{	
\begin{tabular}{l|c|c|c|c|c|l}
\toprule[1.0pt]
Dataset                                               & Dim & \begin{tabular}[c]{@{}c@{}}Input\\ Length\end{tabular}  & \begin{tabular}[c]{@{}c@{}}Prediction\\ Length\end{tabular}    & \begin{tabular}[c]{@{}c@{}}Time\\ Points\end{tabular}  & \begin{tabular}[c]{@{}c@{}}Dataset\\ Splitting\end{tabular}  &  Frequency \\ \midrule[0.5pt]
\begin{tabular}[c]{@{}l@{}}ETTm1\\ ETTm2\end{tabular} & 7   & 336          & \{96,192,336,720\} & 69,680      & 6:2:2        & 15 mins  \\ \midrule[0.5pt]
\begin{tabular}[c]{@{}l@{}}ETTh1\\ ETTh2\end{tabular} & 7   & 336          & \{96,192,336,720\} & 17,420       & 6:2:2        &  Hourly    \\ \midrule[0.5pt]
Electricity                                           & 321 & 336          & \{96,192,336,720\} & 26,304       & 7:1:2        & Hourly     \\ \midrule[0.5pt]
Traffic                                               & 862 & 336          & \{96,192,336,720\} & 17,544       & 7:1:2         & Hourly  \\ \midrule[0.5pt]
Weather                                               & 21  & 336          & \{96,192,336,720\} & 52,696       & 7:1:2        & 10 mins       \\ \midrule[0.5pt]
Exchange                                              & 8   & 336          & \{96,192,336,720\} & 7,588        & 7:1:2        & Daily   
\\ \bottomrule[1.0pt] 
\end{tabular}
}
\end{table}

\begin{table}[ht]
\caption{Other dataset descriptions. The dataset size is organized in (Train,Validation,Test).}
\label{tab_ds3}
\centering
   \renewcommand{\arraystretch}{1.1}
	\resizebox{0.9\linewidth}{!}{	
\begin{tabular}{l|l|c|c|c}
\toprule[1.0pt]
Tasks                        & Dataset              & Dim & \begin{tabular}[c]{@{}c@{}}Series\\ Length\end{tabular} & Dataset Size           
\\ \midrule[0.5pt]
\multirow{4}{*}{\rotatebox{90}{Imputation}}  &  \begin{tabular}[l]{@{}l@{}}ETTm1,ETTm2 \\ (Electricity)\end{tabular}         & 7   & 96            & (34465,11521,11521)      \\ \cmidrule(lr){2-5}
                             &  \begin{tabular}[l]{@{}l@{}}ETTh1,ETTh2\\ (Electricity)\end{tabular}          & 7   & 96            & (8545,2881,2881)         \\ \cmidrule(lr){2-5}
                             & Electricity          & 321 & 96            & (18317,2633,5261)         \\ \cmidrule(lr){2-5}
                             & Weather              & 21  & 96            & (36792,5271,10540)        \\ \midrule[0.5pt]
\multirow{10}{*}{\rotatebox{90}{Classification (UEA)}} &  \begin{tabular}[l]{@{}l@{}}EthanolConcentration\\ (Alcohol Industry)\end{tabular}  & 3   & 1751          & (261,0,263)               \\  \cmidrule(lr){2-5}
                             &\begin{tabular}[l]{@{}l@{}}FaceDetection\\ (Face)\end{tabular}         & 144 & 62            & (5890,0,3524)                  \\ \cmidrule(lr){2-5}
                             & \begin{tabular}[l]{@{}l@{}}Handwriting\\ \end{tabular}\          & 3   & 152           & (150,0.850)                     \\ \cmidrule(lr){2-5}
                             & \begin{tabular}[l]{@{}l@{}}Heartbeat\\ (Heart Beat)\end{tabular}            & 61  & 405           & (204,0,205)                     \\ \cmidrule(lr){2-5}
                             & \begin{tabular}[l]{@{}l@{}}JapaneseVowels\\ (Voice)\end{tabular}       & 12  & 29            & (270,0,370)                \\\cmidrule(lr){2-5}
                             &\begin{tabular}[l]{@{}l@{}}PEMS-SF\\ (Transportation)\end{tabular}               & 963 & 144           & (267,0,173)              \\\cmidrule(lr){2-5}
                             &\begin{tabular}[l]{@{}l@{}}SelfRegulationSCP1\\ (Health)\end{tabular}    & 6   & 896           & (268,0,293)                   \\\cmidrule(lr){2-5}
                             & \begin{tabular}[l]{@{}l@{}}SelfRegulationSCP2\\ (Health)\end{tabular}   & 7   & 1152          & (200,0,180)                    \\\cmidrule(lr){2-5}
                             & \begin{tabular}[l]{@{}l@{}}SpokenArabicDigits\\ (Voice)\end{tabular}   & 13  & 93            & (6599,0,2199)             \\\cmidrule(lr){2-5}
                             & \begin{tabular}[l]{@{}l@{}}UWaveGestureLibrary\\ (Gesture)\end{tabular}  & 3   & 315           & (120,0,320)                   \\ \midrule[0.5pt]
\multirow{5}{*}{\rotatebox{90}{Anomaly Detection}}    & \begin{tabular}[l]{@{}l@{}}SMD\\ (Server Machine
)\end{tabular}                  & 38  & 100           & (566724,141681,708420)       \\\cmidrule(lr){2-5}
                             & \begin{tabular}[l]{@{}l@{}}MSL\\ (Spacecraft)\end{tabular}                  & 55  & 100           & (44653,11664,73729)               \\\cmidrule(lr){2-5}
                             & \begin{tabular}[l]{@{}l@{}}SMAP\\ (Spacecraft)\end{tabular}                 & 25  & 100           & (108146,27037,427617)              \\\cmidrule(lr){2-5}
                             & \begin{tabular}[l]{@{}l@{}}SWaT\\ (Infrastructure)\end{tabular}                 & 51  & 100           & (396000,99000,449919)          \\\cmidrule(lr){2-5}
                             & \begin{tabular}[l]{@{}l@{}}PSM\\ (Server Machine)\end{tabular}                  & 25  & 100           & (105984,26497,87841)          \\ \bottomrule[1.0pt] 
\end{tabular}
}
\end{table}

Timer has curated the Unified Time Series Dataset (UTSD) \cite{liu2024timer}, which encompasses seven domains and includes up to 1 billion time points, organized in a hierarchical four-volume structure to facilitate research and pre-training of large models in the time series field. The Unified Time Series Dataset (UTSD) is meticulously compiled from a combination of publicly accessible online data repositories and empirical data derived from real-world machine operations. All datasets are classified into seven distinct domains by their sources: Energy, Environment, Health, Internet of Things (IoT), Nature, Transportation, and Web, with diverse sampling frequencies. UTSD is structured with hierarchical capacities, namely UTSD-1G, UTSD-2G, UTSD-4G, and UTSD-12G, where each smaller dataset is a subset of the larger ones. A larger subset means greater data difficulty and diversity. For our pre-training tasks, we directly utilized data from the UTSD-12G.

The datasets for downstream tasks include multiple types of test datasets, mainly covering long-term time series forecasting, imputation, classification, and anomaly detection tasks. The datasets for long-term time series forecasting include Electricity, Weather, Traffic, and the ETT family (comprising ETTh1, ETTh2, ETTm1, and ETTm2), which are utilized to evaluate long-term performance in scenarios such as energy consumption, weather forecasting, and traffic flow analysis. The imputation task datasets include Electricity, Weather, and the ETT family, focusing on solving missing data issues. The time series classification datasets are sourced from the UEA database and encompass tasks from various fields, such as EthanolConcentration (alcohol industry), Heartbeat (health monitoring), JapaneseVowels (speech recognition), PEMS-SF (traffic and transportation), and more. Anomaly detection tasks are based on five standard time series anomaly detection datasets, including SMD (server machines), MSL and SMAP (spacecraft), SWaT (infrastructure systems), and PSM (server machines). These datasets span diverse application domains, including energy, transportation, healthcare, industrial control, and space science, providing comprehensive test scenarios for various tasks.

\subsection{Pre-training Configuration}

TimeHAT is instantiated with a total of $N=8$ attention layers, comprising $M=3$ prefix-constrained BiA layers at the bottom and 5 decoupled HA layers at the top stages. We set the hidden dimension $d_{model} = 1024$, feed-forward dimension $d_{ff} = 2048$, and the number of attention heads $H=16$. For temporal tokenization, the input context length is fixed at $T=1440$ with a patch size of $P=96$. The resulting model contains approximately 67M trainable parameters.

We pre-train the model on the UTSD-12G dataset using UPPMS, which Reconstructive Sampling objective with a 40\% masking ratio. The optimization is performed using AdamW with a weight decay of 0.01. Given the extreme sparsity of the reconstructive objective and the massive scale of the pre-training corpus, we employ a large global batch size of 10,240 to stabilize the gradient variance. Correspondingly, rather than relying on computationally expensive empirical grid search, we determine the peak learning rate via the Square Root Scaling Rule tailored for adaptive optimizers. Adopting the widely accepted empirical baseline of $LR_{base} = 1 \times 10^{-4}$ for a standard batch size of $BS_{base} = 256$, the scaled learning rate is computed as:
$$
    LR_{scaled} = LR_{base} \times \sqrt{\frac{BS_{large}}{BS_{base}}} \approx 6.3 \times 10^{-4}
$$
While the scaling rule provides a theoretical baseline, masked reconstructive objectives typically suffer from gradient sparsity during the early stages. To compensate for this sparsity and the rapid descent nature of the cosine schedule, we empirically anchor the final peak learning rate slightly higher at $8 \times 10^{-4}$. To prevent early gradient shock under this elevated learning rate, we apply a strict linear warmup of 4,000 steps, which eventually decays to a minimum of 10\% of the peak.

The maximum training duration is set to 20 epochs. Given the extreme scale of the pre-training corpus, we adopt a high-frequency, step-based evaluation strategy rather than epoch-based validation. Specifically, the validation loss is evaluated every 5,000 steps. Training is halted via an early stopping mechanism if the validation loss does not decrease for 10 consecutive evaluations. All pre-training runs are executed using Distributed Data Parallel (DDP) across four NVIDIA RTX 5090 (32GB) GPUs.

\begin{table*}[h]
    \centering
    \caption{Forecasting performance with embedding and encoder transfer.
    $P$ denotes the embedding patch size.
    Scratch: training from scratch; Patch: patch-level masking;
    Point: point-level masking.}
    \label{tab:embedding_encoder_transfer}
    \begin{tabular}{c|cc|cc|cc|cc|cc|cc}
        \toprule
        \multirow{3}{*}{Dataset}
        & \multicolumn{6}{c|}{$P=1$}
        & \multicolumn{6}{c}{$P=16$} \\
        \cmidrule(lr){2-7}\cmidrule(lr){8-13}
        & \multicolumn{2}{c|}{Scratch}
        & \multicolumn{2}{c|}{Patch}
        & \multicolumn{2}{c|}{Point}
        & \multicolumn{2}{c|}{Scratch}
        & \multicolumn{2}{c|}{Patch}
        & \multicolumn{2}{c}{Point} \\
        & MSE & MAE & MSE & MAE & MSE & MAE
        & MSE & MAE & MSE & MAE & MSE & MAE \\
        \midrule
        ETTh1
        & \textbf{0.465} & \textbf{0.466}
        & 0.465 & 0.468
        & 0.466 & 0.469
        & 0.458 & 0.453
        & 0.447 & 0.445
        & \textbf{0.444} & \textbf{0.444} \\
        ETTh2
        & 0.406 & \textbf{0.422}
        & 0.405 & 0.424
        & \textbf{0.401} & 0.422
        & 0.386 & 0.410
        & \textbf{0.385} & \textbf{0.410}
        & 0.387 & 0.410 \\
        ETTm1
        & \textbf{0.403} & \textbf{0.415}
        & 0.408 & 0.419
        & 0.403 & 0.417
        & 0.393 & 0.406
        & 0.383 & 0.402
        & \textbf{0.383} & \textbf{0.401} \\
        ETTm2
        & \textbf{0.294} & \textbf{0.339}
        & 0.306 & 0.349
        & 0.298 & 0.343
        & 0.290 & 0.336
        & 0.291 & 0.337
        & \textbf{0.289} & \textbf{0.334} \\
        Weather
        & 0.257 & 0.281
        & 0.255 & 0.281
        & \textbf{0.255} & \textbf{0.280}
        & 0.255 & 0.277
        & \textbf{0.253} & \textbf{0.276}
        & 0.254 & 0.276 \\
        \bottomrule
    \end{tabular}%
\end{table*}

\begin{table}[h]
    \centering
    \caption{Forecasting performance with encoder-only transfer.
    All downstream models use a newly initialized $P=16$ embedding
    and prediction head. $P$ denotes the embedding patch size.
    Patch and Point denote patch-level and point-level masking,
    respectively.}
    \label{tab:encoder_only_transfer}
    \resizebox{1\linewidth}{!}{	
    \begin{tabular}{c|cc|cc|cc|cc}
        \toprule
        \multirow{2}{*}{Dataset}
        & \multicolumn{2}{c|}{\makecell{Patch\\$P=1$}}
        & \multicolumn{2}{c|}{\makecell{Patch\\$P=16$}}
        & \multicolumn{2}{c|}{\makecell{Point\\$P=1$}}
        & \multicolumn{2}{c}{\makecell{Point\\$P=16$}} \\
        & MSE & MAE & MSE & MAE
        & MSE & MAE & MSE & MAE \\
        \midrule
        ETTh1
        & 0.450 & 0.449
        & 0.448 & 0.445
        & 0.456 & 0.452
        & \textbf{0.447} & \textbf{0.445} \\
        ETTh2
        & 0.389 & 0.411
        & \textbf{0.385} & \textbf{0.410}
        & 0.387 & 0.411
        & 0.387 & 0.410 \\
        ETTm1
        & 0.392 & 0.405
        & \textbf{0.386} & \textbf{0.403}
        & 0.393 & 0.407
        & 0.387 & 0.404 \\
        ETTm2
        & 0.288 & 0.334
        & 0.288 & 0.333
        & 0.288 & 0.334
        & \textbf{0.286} & \textbf{0.332} \\
        Weather
        & 0.254 & 0.277
        & 0.254 & 0.277
        & 0.254 & 0.277
        & \textbf{0.253} & \textbf{0.276} \\
        \bottomrule
    \end{tabular}
    }
\end{table}

\subsection{Downstream Tasks}

Unlike conventional paradigms that discard the pre-training head, we retain both the pre-trained TimeHAT backbone and its projection head for sequence-to-sequence downstream tasks (i.e., forecasting, imputation, and anomaly detection). This unified architecture maximizes the direct transferability of the learned reconstructive representations. An additional classification head is appended exclusively for the classification task. 

To ensure stable adaptation and mitigate the catastrophic forgetting of pre-trained representations, we employ a differential learning rate strategy during fine-tuning. Specifically, the optimizer is re-initialized with a conservative peak learning rate for the pre-trained backbone ($LR_{\text{backbone}} \in [1 \times 10^{-5}, 3 \times 10^{-5}]$), while a relatively higher learning rate is allocated to the task-specific projection and classification heads ($LR_{\text{head}} \in [5 \times 10^{-5}, 1 \times 10^{-4}]$). This optimization is paired with a short linear warmup comprising 5\% of the total steps. The fine-tuning process spans a maximum of 20 epochs, governed by an early stopping mechanism with a patience of 5 epochs.

\section{Effect of Masking Granularity and Embedding Size}

We compare point-level and patch-level masking with embedding patch sizes
$P=1$ and $P=16$ under two transfer protocols: transferring both the
embedding and encoder, and transferring only the encoder to a common
$P=16$ downstream embedding. Results are averaged over four prediction
horizons and three fine-tuning seeds.
Tables~\ref{tab:embedding_encoder_transfer} and
\ref{tab:encoder_only_transfer} show that patch embeddings ($P=16$)
generally yield better forecasting performance than point-wise embeddings
($P=1$), particularly under encoder-only transfer. Notably, point-level
masking remains effective with $P=16$, achieving the best results on
several datasets. These findings indicate that fine-grained masking
supervision can be combined with patch-based representations, without
requiring point-wise tokenization.

\begin{figure}[h]
    \centering
    \includegraphics[scale=0.24]{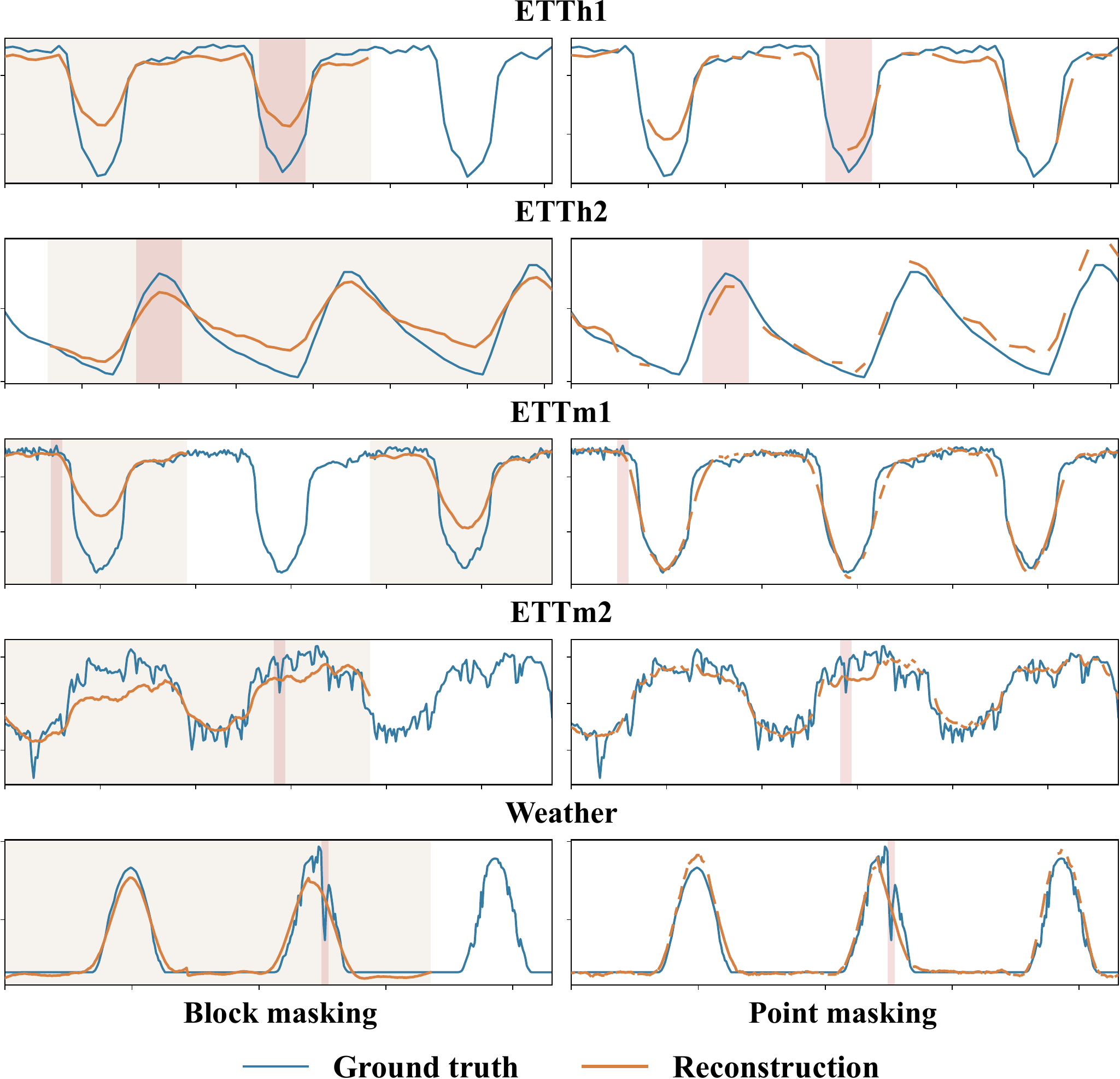}
    \caption{Qualitative reconstruction examples under block and point
    masking on five datasets. Each row shows the same selected sequence
    and vertical scale.}
\label{fig:masking_visualization}
\end{figure}

We compare block-level and point-level masking on five datasets using
matched masking ratios. Figure~\ref{fig:masking_visualization} shows that
both approaches recover the major temporal patterns, but block masking
often underestimates peak and trough amplitudes and misses fine-grained
variations, particularly on ETTh1 and ETTm1. Point masking generally
produces reconstructions closer to the ground truth, with clearer
preservation of local details.

\end{document}